%% file: main.tex
\documentclass[twocolumn]{autart}    

\usepackage{graphicx}          
\usepackage{url}
\usepackage{amsmath}
\usepackage{amssymb}
\usepackage{wasysym}
\usepackage{enumerate}
\usepackage{algorithm} 
\usepackage[noend]{algpseudocode}
\usepackage{booktabs}
\usepackage{multirow}
\usepackage{acro}
\usepackage{xcolor}
\acsetup{patch/maketitle=false}
\input{./acronyms} 
\makeatletter
\let\cl@part\@empty 
\makeatother
\usepackage{cleveref}
\crefname{equation}{Eq.}{Eqs.} 
\Crefname{equation}{Equation}{Equations} 
\Crefname{figure}{Fig.}{Figs.} 
\newtheorem{theorem}{Theorem}

\DeclareMathOperator*{\argmax}{argmax}
\DeclareMathOperator*{\argmin}{argmin}

\journal{Computers \& Chemical Engineering}

\begin{document}

\begin{frontmatter}

\title{Model-Free Learning of Model-Based Policies: A Policy Optimization Perspective on Safe, Interpretable, and Sample-Efficient Control
\thanksref{footnoteinfo}} 

\thanks[footnoteinfo]{This paper was not presented at any IFAC meeting. Corresponding author: A. Mesbah.}

\author[UCB]{Thomas Banker}~and
\author[UCB]{Ali Mesbah}\ead{mesbah@berkeley.edu}

\address[UCB]{Department of Chemical and Biomolecular Engineering, University of California, Berkeley, CA 94720, USA.}


\begin{abstract}                          

Controlling engineered systems requires sequential decisions that optimize long-term performance and satisfy operating constraints despite uncertainty in the system and its environment.
Policy optimization addresses such decision-making problems by parameterizing a feedback policy and adjusting its parameters to optimize a closed-loop performance objective.
Policy optimization methods are conventionally divided into model-based and model-free approaches.
Model-based methods derive the policy from a system model, exploiting prior knowledge but losing performance and guarantees on constraint satisfaction under model inaccuracies.
Model-free methods learn the policy directly from data, avoiding model bias but requiring many samples and yielding black-box policies without constraint guarantees.
This perspective paper argues for combining model-based and model-free approaches through model-free learning of model-based policies: policies defined by parametric optimization problems, such as model predictive control, whose dynamics, cost, and constraint parameters are tuned directly from system data.
The prior model provides the policy with interpretable structure, modularity, and a basis for constraint satisfaction, while model-free updates correct for model inaccuracies to improve closed-loop performance.
Related ideas have been developed separately in control, process systems engineering, and machine learning with distinct terminology and assumptions, obscuring shared principles.
Through this unifying lens, we frame Bayesian optimization, reinforcement learning, and behavioral cloning as distinct routes to model-free learning of model-based policies, each with its own strengths and limitations. 
We then identify research directions for this paradigm, including learning schemes that preserve control-theoretic safety guarantees, scalable computation for high-dimensional systems, and learning beyond Markov decision processes.
By connecting established model-free learning methods with the structure of model-based policies, this perspective aims to lay a common foundation for sample-efficient, safe, and interpretable decision-making under uncertainty.
\end{abstract}

\end{frontmatter}

\acresetall

\section{Introduction} \label{sec: 1}

As engineered systems grow more complex, interconnected, and data-rich, control policies operating these systems must increasingly be optimized from system data rather than designed from prior knowledge alone.
Data-driven policies already operate chemical processes \cite{gao2024reinforcement,park2025real,gil2026reinforcement}, regulate energy use in buildings \cite{azuatalam2020reinforcement,wang2020reinforcement}, stabilize plasma in fusion reactors \cite{degrave2022magnetic}, and control robots from camera images \cite{levine2016end-to-end}.
Optimizing policies from data extends beyond control, from molecular design \cite{popova2018deep} to multi-step reasoning in large language models \cite{deepseekai2025deepseekr1}.
Despite their different domains, these decision-making problems share a common formulation: a policy is selected from a parameterized class, whose structure can reflect available prior knowledge, by optimizing a performance function that encodes the system objectives and constraints.

System properties, however, restrict which optimization methods can be applied.
A system may be accessible only through real-world interactions; its model may be uncertain, incomplete, or unknown; or its state and action spaces may be too large for exact solution methods, such as value and policy iteration \cite{bertsekas1995dynamic} or solving the Hamilton–Jacobi–Bellman equation \cite{kalman1963theory}.
Under these limitations, the performance of a candidate policy---even one informed by prior knowledge---generally cannot be evaluated exactly, much less the performance gradient.
Both must instead be estimated from samples, such as noisy evaluations of the performance function or its gradients.
Policy optimization refers to the class of methods that update the policy parameters directly from sample-based estimates \cite{hu2023toward}.
Policy optimization methods differ mainly in the type of data available and in how the data guide the policy parameter updates.
\Ac{RL} estimates performance gradients or value functions from rewards observed during system interaction \cite{sutton2018reinforcement}.
\ac{BO} fits a probabilistic surrogate of performance as a function of policy parameters from costly evaluations of the performance function and uses the surrogate to select policy parameters \cite{shahriari2015taking}.
\Ac{BC} replaces the performance function with a measure of discrepancy between the policy's decisions and an expert's demonstrations, improving performance only indirectly, to the extent that the expert performs well \cite{osa2018algorithmic}.

Policy optimization methods are traditionally categorized by whether they use a model of the system dynamics.
Model-free methods use no dynamics model and update the policy directly from system samples \cite{sutton2018reinforcement}.
Model-free methods scale to high-dimensional state and action spaces with limited assumptions on the system \cite{mnih2015human-level,lillicrap2016continuous}, but often require many interactions and, when paired with black-box policy classes such as neural networks, generally lack guarantees of state constraint satisfaction during learning or deployment \cite{mania2018simple,recht2019tour,brunke2022safe}.
Model-based methods seek to address these limitations by incorporating a dynamics model into policy optimization \cite{sutton1991dyna,berkenkamp2017safe,moerland2022model-based}.
Model inaccuracies, however, can degrade closed-loop performance and lead to constraint violations \cite{gevers2005identification,lambert2020objective,hewing2020learning-based}.
How best to leverage imperfect models in policy optimization remains largely an open question \cite{wang2019benchmarking,moerland2022model-based}.

In addressing this question, it is useful to distinguish two roles a model can play in policy optimization, both commonly described as \emph{model-based}.
In one role, the model acts as a surrogate of the system, generating predicted samples, gradients, or performance estimates that accelerate the optimization, as is common in model-based \ac{RL} \cite{janner2021when,heess2015learning,feinberg2018model-based}.
Alternatively, the model can form part of the policy itself, as in \ac{MPC}, where the policy is defined by an optimization problem over trajectories predicted by the dynamics model, subject to a parameterized cost and constraints \cite{rawlings2017model}.
Such model-based policies lie at the opposite end of a spectrum of policy classes from generic parameterizations, such as neural networks, which assume little about the system but offer no structure to interpret or constrain the resulting decisions.
Separating these roles reveals that, when the model belongs to the policy rather than the learning process, a model-based policy parameterization is compatible with model-free learning.

This paper presents a policy optimization perspective on model-free learning of model-based policies.
We treat the dynamics model, cost, and constraints of a model-based policy as learnable policy components rather than fixed design choices. 
Updating these components through policy optimization on samples of the true system makes the learning model-free; closed-loop performance can improve despite model inaccuracies, while the optimization structure acts as an inductive bias.
Compared with neural network policies, a model-based policy typically involves fewer and physically meaningful parameters, yields interpretable quantities such as predicted trajectories and active constraints, and provides a basis for constraint satisfaction.
We characterize these advantages together with the practical challenges they bring, including the computational cost of solving and differentiating an optimization problem at every decision, and the potential loss of nominal guarantees as policy parameters are learned.

Within this perspective, we then discuss how \ac{BO}, \ac{RL}, and \ac{BC} serve as three routes to learning policies, and how the structure of model-based policies shapes each route.
\Ac{BO} requires no derivatives of the policy and can therefore treat a model-based policy as a black box, but it is typically effective only for low-dimensional policy parameterizations \cite{paulson2023tutorial}.
\Ac{RL} and \ac{BC} scale more readily, but for model-based policies, they additionally require derivatives of the optimal solution with respect to policy parameters \cite{gros2020data-driven,amos2018differentiable}, which can be costly to obtain repeatedly via \ac{IFT} \cite{dontchev2009implicit,xu2023revisiting}.
Viewing the three methods through this shared lens exposes common bottlenecks that currently limit practical adoption and points to promising directions for future research: learning schemes that preserve control-theoretic safety guarantees throughout learning; scalable computation of optimization-based policies and their derivatives for high-dimensional systems; and extensions beyond Markov decision processes to partially observed, multi-agent, and nonstationary settings. 

The remainder of the paper is organized as follows. \Cref{sec: 2} formulates policy optimization in \iac{MDP} and reviews the main classes of solution approaches.
\Cref{sec: 3} introduces model-based policies, describes their design through policy optimization, and discusses their benefits relative to generic function approximators.
\Cref{sec: 4,sec: 5,sec: 6} survey model-free approaches to learning model-based policies, covering \ac{BO}, policy gradient \ac{RL}, and methods that learn without system interaction, namely \ac{BC} and offline \ac{RL}.
Each of these sections closes with future research opportunities.

\section{From Dynamic Programming to Policy Optimization} \label{sec: 2}

\subsection{Markov Decision Processes} \label{sec: 2.1}

An \ac{MDP} formalizes sequential decision-making under uncertainty, in which each decision influences, but does not fully determine, the subsequent evolution of the system.
The uncertainty in this evolution is described probabilistically: at each discrete time step~$t$, the system is described by its state $s_t \in \mathcal{S} \subseteq \mathbb{R}^{n}$, and its next state follows a probability distribution conditioned on the current state and decision.
Starting from some initial state distribution $p(s_{0})$, the decision-making agent selects an input or action $a_{t} \in \mathbb{R}^{m}$ from the action space $\mathcal{A}$ according to its \emph{control policy} $\pi \in \Pi$.
This policy may select actions in a stochastic manner $a \sim \pi(a|s)$, assigning each action a probability, or deterministically $a = \pi(s)$ for a given state.
Inputting $a_{t}$ to the system, the successor state $s_{t+1}$ at the next time $t+1$ and reward  $r_{t} \in \mathbb{R}$ received by the agent are governed by the probabilistic state transition dynamics $s_{t+1} \sim p(s_{t+1} | s_{t}, a_{t})$ and reward function $r_{t} = r(s_{t},a_{t})$, respectively.
Here, we assume the dynamics to be time-invariant, satisfying the Markov property:
\[
    p(s_{t+1}|s_{0},a_{0},\ldots,s_{t},a_{t}) = p(s_{t+1}|s_{t},a_{t}),
\]
stating that the future evolution of the stochastic process is independent of its history given the current state-action pair $s_{t},a_{t}$. 
By repeatedly selecting actions according to the policy $\pi$, a trajectory of state-action pairs $\tau = (s_{0},a_{0},s_{1},a_{1},s_{2},\ldots)$ is generated with the corresponding reward sequence $(r_{0},r_{1},\ldots)$.

The policy's performance is evaluated by a \emph{performance function} $J(\pi): \Pi \rightarrow \mathbb{R}$, formulated as the expected cumulative discounted rewards received across trajectories:
\begin{equation} \label{eqn: performance function}
    J(\pi) = \mathbb{E}_{\pi} \Biggl\{ \sum_{t=0}^{\infty} \gamma^{t} r(s_{t}, a_{t}) \Biggr\},
\end{equation}
where $\gamma \in (0,1)$ is a discount factor and the expected value $\mathbb{E}_{\pi}\{ \cdot \}$ is taken over the closed-loop trajectories resulting from a stochastic policy $\pi(a|s)$ and probabilistic state transition dynamics $p(s_{t+1} | s_{t}, a_{t})$, starting from $s_{0} \sim p(s_{0})$.
Formally, an optimal control policy for \iac{MDP} maximizes the performance function, which can be obtained by solving a mathematical optimization:
\begin{equation} \label{eqn: optimal policy}
        \pi^{\star} = \argmax_{\pi \in \Pi} ~ J(\pi).
\end{equation}

\subsection{Value Functions and Dynamic Programming}

Solving for the optimal policy can be facilitated by introducing the state-value function and action-value function, commonly referred to as the value function and $Q$-function, respectively \cite{bertsekas1996neuro-dynamic}.
The value function $V^{\pi}(s) : \mathbb{R}^{n} \rightarrow \mathbb{R}$ shares similar structure to \cref{eqn: performance function}, representing the expected rewards from following policy $\pi$ but conditioned on a particular initial state $s$:
\begin{equation} \label{eqn: value function}
    V^{\pi}(s) = \mathbb{E}_{\pi} \Biggl\{ \sum_{t=0}^{\infty} \gamma^{t} r(s_t,a_t) \Bigg| s_0 = s \Biggr\}.
\end{equation}
Assuming the agent selects some initial action $a$, the $Q$-function $ Q^{\pi}(s,a): \mathbb{R}^{n} \times \mathbb{R}^{m} \rightarrow \mathbb{R}$ can be similarly defined as:
\begin{equation} \label{eqn: Q-function}
    Q^{\pi}(s,a) = \mathbb{E}_{\pi} \Biggl\{ \sum_{t=0}^{\infty} \gamma^{t} r(s_t,a_t) \Bigg| s_0 = s, a_0 = a \Biggr\}.
\end{equation}
While \cref{eqn: value function,eqn: Q-function} encode similar information, the $Q$-function allows for reasoning about the long-term consequences of actions without explicitly using the dynamics.
Furthermore, the value function can be expressed in terms of the $Q$-function (and vice versa) with the Bellman recursion:
\begin{subequations}\label{eqn: value and Q-function relations}
    \begin{align}
        V^{\pi}(s) =&~\mathbb{E}_{a \sim \pi(a|s)} \bigl\{ Q^{\pi}(s,a) \big\},\\
        Q^{\pi}(s,a) =&~r(s,a) + \gamma \mathbb{E}_{s' \sim p(s'|s,a)} \big\{ V^{\pi}(s') \big\}.
    \end{align} 
\end{subequations}
With this relationship in mind, we focus our discussion on the $Q$-function.

The optimal $Q$-function $Q^{\star}(s,a)$ encodes the maximum expected future rewards for each state-action pair
\begin{equation} \label{eqn: optimal Q-function}
    Q^{\star}(s,a) = \sup_{\pi \in \Pi} Q^{\pi}(s,a), \quad \forall (s,a) \in \mathcal{S} \times \mathcal{A}.
\end{equation}
This optimal $Q$-function uniquely satisfies the Bellman optimality equation \cite{bellman1957dynamic}:
\begin{equation} \label{eqn: bellman optimality equation}
    Q^{\star}(s,a) = r(s,a) + \gamma\mathbb{E}_{s' \sim p(s'|s,a)} \biggl\{ \max_{a'\in\mathcal{A}} Q^{\star}(s',a') \biggr\}.
\end{equation}
The Bellman optimality equation can be regarded as a fixed-point equation, i.e., $Q^{\star}(s,a) = \mathcal{T}Q^{\star}(s,a)$ where $\mathcal{T}$ is a contraction mapping \cite{lewis2012reinforcement}.
Thus, not only does \cref{eqn: bellman optimality equation} describe a global condition for optimality, but it also provides a path for obtaining the optimal $Q$-function with iterative fixed-point methods.

Provided one can obtain $Q^{\star}(s,a)$, an optimal policy can be recovered by ``greedily'' taking actions via maximizing the $Q$-function
\begin{equation} \label{eqn: optimal greedy policy}
    \pi^{\star}(s) = \argmax_{a} Q^{\star}(s,a).
\end{equation}
While described as greedy, such a policy is globally optimal because $Q^{\star}$ accounts for rewards in the potentially infinitely-distant future.
However, its evaluation only requires a state-action pair, making the future value locally and immediately available.
Hence, a greedy one-step-ahead search of the optimal $Q$-function yields a globally optimal policy.

The aforementioned \ac{DP} approach can solve for the optimal $Q$-function through the Bellman equation to obtain the optimal policy $\pi^\star$, but while theoretically powerful, such an approach is challenging to use in practice.
The \ac{DP} solution requires complete knowledge of the state transition dynamics, and even with complete knowledge, recursive solution of the Bellman equation can become intractable due to the ``curse of dimensionality.''
This ``curse'' refers to the exponential rise in computation time required to approximate the solution to \iac{MDP}.
As the number of discretized states grows exponentially with the state dimension, the multi-variable integrals of \cref{eqn: bellman optimality equation} become difficult to evaluate, making \ac{DP} computationally intractable.
           
\subsection{Approximate Solutions to DP via Model-free Learning} \label{sec: 2.3}

The intractability of \ac{DP} motivates approximate solution methods that can accommodate large state-action spaces while circumventing the intractable integrals that also require exact knowledge of the dynamics.
\emph{Model-free learning} methods address both limitations via two key approximations.
First, the policy and $Q$-function are represented by function approximators $\pi_\theta$ and $Q_\phi$, with learnable parameters $\theta$ and $\phi$, which avoids discretizing the state-action space.
Second, the expectations are estimated from sampled transitions of the true system rather than computed with a known model of its dynamics; this is what makes learning model-free.
Model-free learning thereby avoids the bias that inaccurate knowledge of the dynamics may introduce, though typically at the cost of requiring many samples.
Critically, model-free learning is a property of the learning signal, not of the function approximator.
A policy (or value) function approximator may itself contain a model, as discussed in \Cref{sec: 3}, and still be learned in a model-free manner.

Model-free learning methods are commonly distinguished by which function approximators they learn:
\begin{enumerate}[(i)]
    \item \emph{value-based} methods, which learn a value function approximator that best satisfies the Bellman optimality equation, i.e., \cref{eqn: bellman optimality equation}, deriving a policy then after; and
    \item \emph{policy-based} methods, which directly design a policy function approximator to maximize performance, i.e., \cref{eqn: performance function}.
\end{enumerate}
Policy-based methods are representative of policy optimization, the focus of this paper.
Value-based methods instead maintain a closer connection to \ac{DP}, learning through the Bellman equation, but remain relevant to policy optimization.
For example, a learned value function can serve as a critic that guides policy updates in actor-critic methods \cite{grondman2012survey}, which combine policy- and value-based approximations.

\emph{Value-based methods.}\quad As suggested by \cref{eqn: optimal greedy policy}, one way to (approximately) recover an optimal policy that maximizes \cref{eqn: performance function} is by first learning an approximation of $Q^{\star}$.
A value-based approach draws upon \ac{DP} to iteratively update a $Q$-function approximator such that its $Q$-values best satisfy the Bellman optimality equation \cite{watkins1992q-learning}.
Value-based approaches generally solve \ac{TD} learning problems of the form:
\begin{equation} \label{eqn: mean squared bellman error minimization}
    \phi^{\star} = \argmin_{\phi} \mathbb{E}_{(s,a,r,s')}\bigg\{ \Big( Q_{\phi}(s,a) - q \Big)^2 \bigg\},
\end{equation}
where $Q_{\phi}$ denotes a parametric $Q$-function approximator with learnable parameters $\phi$ and $q \in \mathbb{R}$ is the temporal difference target. The choice of the temporal difference target depends on the \ac{RL} algorithm.
Commonly, $q$ is designed to reflect the right-hand side of \cref{eqn: bellman optimality equation} as 
\[
q = r + \gamma \max_{a'} Q_{\phi}(s',a').
\]
Accordingly, the expectation in \cref{eqn: mean squared bellman error minimization} is taken with respect to tuples of the form $(s,a,r,s')$, which is approximated using samples of system interactions collected under sufficiently exciting inputs.
With this target, \cref{eqn: mean squared bellman error minimization} is an example of off-policy learning, which learns the optimal $Q$-function for a ``greedy'' policy---like that of \cref{eqn: optimal greedy policy}---not necessarily the same as the exploration policy that gathered the samples.
The optimal parameters $\phi^{\star}$, which solve \cref{eqn: mean squared bellman error minimization}, can be seen as minimizing the error between the $Q$-function and its one-step recursion.

Upon solving \cref{eqn: mean squared bellman error minimization}, the decision-making agent designs a policy as an approximation to the optimal policy $\pi^{\star}$.
This policy may be optimization-based, computing the action that maximizes $Q_\phi$ at each state similar to \cref{eqn: optimal greedy policy} \cite{mnih2015human-level,banker2026soft}, or represented by a generic function approximator, designed to approximately maximize $Q_\phi$ at each state.
The former represents a strictly value-based approach, while the latter is an example of an actor-critic algorithm for policy optimization.

\emph{Policy-based methods.}\quad While value-based methods first learn an approximation of $Q^{\star}$ and then derive the policy from it, policy-based approaches can learn a policy directly from sampled system interactions.
These approaches commonly consider a parametric policy representation $\pi_{\theta}$, with learnable parameters $\theta$.
The parameters are designed to maximize \cref{eqn: performance function}---or a surrogate objective such as agreement with expert demonstrations---casting policy optimization as\footnote{In practice, there may be multiple performance functions $\{ J_{j} \}_{j=0}^{N_{J}-1}$ based on which the control policy must be designed \cite{makrygiorgos2022performance-oriented}.
For example, these measures may correspond to reference tracking error, nonlinear economic costs, and violation of quality or safety-critical constraints.
For simplicity of exposition, we focus our discussion on a single-objective, unconstrained problem in \cref{eqn: policy search}.}:     
\begin{equation} \label{eqn: policy search}
    \theta^{\star} = \argmax_{\theta} ~ J(\pi_{\theta}).
\end{equation}

With the policy optimization problem \cref{eqn: policy search} as our focus, both the tractability of the problem and the properties of its solution hinge on two considerations: the information available about the performance function, and the policy parameterization. 
Because the performance function generally has no closed-form expression, neither its value nor its gradient with respect to the policy parameters can be computed directly. 
Instead, both must be estimated from closed-loop trajectories, which may be noisy and expensive to obtain \cite{recht2019tour}. 
The type of information available, whether zeroth-order evaluations alone and/or gradient estimates, governs which solution methods are applicable and how efficiently they learn; see \Cref{sec: 3.3} for an overview of solution methods.

The policy parameterization is equally consequential.
For example, gradient estimates require derivatives of the policy with respect to its parameters; the parameterization dictates whether these derivatives exist and how they are computed.
A \ac{NN} policy is differentiable by construction, with derivatives obtained by backpropagation \cite{rumelhart1986learning}.
On the other hand, a policy defined by an optimization problem is differentiable only where its optimal solution varies smoothly with the parameters, a property that can fail when the set of active constraints changes \cite{agrawal2020learning}.
Beyond the learning process, the parameterization also determines the properties of the resulting policy, including its closed-loop performance, interpretability, and capacity to satisfy constraints.
Expressive policy classes such as \acp{NN} impose few restrictions, but model-free learning over such classes can require impractically many samples and generally lacks safety guarantees during learning \cite{mania2018simple,recht2019tour,brunke2022safe}.
Incorporating a model can mitigate both limitations, to an extent that depends on the role the model plays.
One such role is for the model to form part of the policy itself.
The next section defines the notion of a model-based policy and describes how it can be learned in a model-free manner.

\section{Model-Free Learning of Model-Based Policies} \label{sec: 3}

\begin{figure*}[t]
    \centering
    \includegraphics[width=\textwidth]{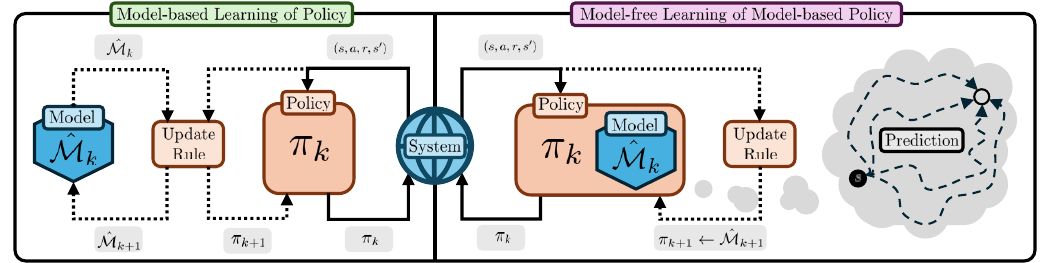} 
    \caption{
    Two roles of a system model in policy optimization. 
    Left: As a surrogate of the system, the model guides learning, for example by supplying the update rule with model-generated samples, targets, or losses.
    The model can still be updated using data from the true system, but separately from the policy, and is generally not used for online decision-making.
    Right: In a model-based policy, the model forms part of the policy, which computes actions online by optimizing over model predictions.
    The effects of the model on decision-making are accounted for in the update rule, which adapts the model along with the rest of the policy using samples of the true system to optimize performance.
    }
    \label{fig: model-based RL}
\end{figure*}

We distinguish two roles a model can play in policy optimization, as illustrated in \Cref{fig: model-based RL}.
In one role, the model acts as a surrogate of the system that guides learning, so that learning is no longer model-free in the sense of the solution methods discussed in \Cref{sec: 2.3}.
Dyna-style methods generate synthetic samples with the model for use in an otherwise model-free learning algorithm \cite{sutton1991dyna}, e.g., to approximate the expectation in \cref{eqn: mean squared bellman error minimization}, but can become brittle when they rely heavily on model-generated experience \cite{aminmansour2024mitigating,barkley2025stealing}.
Value expansion methods use short model rollouts to construct value targets for \cref{eqn: mean squared bellman error minimization} within actor-critic algorithms \cite{feinberg2018model-based}.
However, model errors accumulate with the rollout horizon, and even with a perfect dynamics model, the gains over model-free learning can diminish \cite{palenicek2026diminishing}.
Other methods estimate the gradient of \cref{eqn: performance function} by differentiating through model rollouts \cite{heess2015learning,byravan2020imagined}, which compounds prediction error with exploding or vanishing gradients in backpropagation over long horizons \cite{suh2022differentiable,palenicek2026diminishing}.
In all three cases, the model is fit to minimize prediction error rather than to maximize closed-loop performance, a discrepancy known as objective mismatch \cite{lambert2020objective}. 
Moreover, the model is absent from the deployed policy.
Hence, constraints cannot be enforced through online prediction, and closed-loop performance and constraint satisfaction depend largely on how well the learned policy generalizes.

Alternatively, the model can be embedded in the policy, which computes actions online by optimizing over model predictions subject to constraints.
In this case, the model defines the policy class rather than the learning procedure, yielding a \emph{model-based policy}.
Planning-based \ac{RL} methods also use the model online in a similar way \cite{chua2018deep}, but typically learn the model by minimizing prediction error.
Thus, the objective mismatch of the first role persists, though it can be mitigated when model learning incorporates performance signals \cite{hansen2022temporal}.
Treating the dynamics model, cost, and constraints as learnable policy components instead allows policy optimization to tune them from samples of the true system for closed-loop performance, which addresses the objective mismatch.
This section discusses model-based policies, delineates their advantages relative to \ac{NN} parameterizations, and outlines how they are learned in a model-free manner.

\subsection{Choice of Policy Class} \label{sec: 3.1}

We distinguish two broad classes of policy function approximators:
\begin{enumerate}[(i)]
    \item \emph{\ac{NN} policies} that represent the policy as \iac{NN} \cite{lillicrap2016continuous,schulman2017proximal,haarnoja2018soft}; and
    \item \emph{model-based policies} that represent the policy in terms of adaptable models for the dynamics, cost, and potentially constraints.
\end{enumerate}
We argue the latter provides an inductive bias toward safe, interpretable, and sample-efficient control.

\textbf{\ac{NN} policies.}\quad Such policies have demonstrated empirical success in applications with high-dimensional, continuous state-action spaces, nonlinear dynamics, changing goals, and long horizons \cite{levine2016end-to-end,degrave2022magnetic,andrychowicz2017hindsight,hafner2025mastering}.
This success stems, in part, from two properties of \ac{NN} representations: expressivity and explicit form.
As for expressivity, the universal approximation theorem states that, under mild conditions on the activation function, a sufficiently large \ac{NN} can approximate any continuous function on a compact domain \cite{cybenko1989approximation,hornik1989multilayer}.
In principle, \iac{NN} can serve as a single global approximator of the solution to \cref{eqn: policy search}, even over high-dimensional state spaces, provided the approximator is sufficiently rich in parameterization.
Furthermore, the outputs of \iac{NN} can be explicitly expressed in terms of the input and its learnable parameters.
Actions are therefore obtained with a single forward pass, enabling fast policy evaluation, while policy gradients with respect to the parameters are obtained cheaply by backpropagation.
Consequently, \ac{NN} policies are naturally suited to gradient-based policy learning schemes that can scale with data and network size, which has driven the development of numerous deep \ac{RL} algorithms \cite{lillicrap2016continuous,schulman2017proximal,haarnoja2018soft}.

Despite these advantages, \ac{NN} policies have three key limitations.
First, their generic structure encodes no prior knowledge of the system's dynamics, cost, or constraints.
Combined with heuristically chosen architectures and random initialization, this lack of structure means that policy learning typically requires many samples and gradient updates to reach satisfactory performance \cite{duan2016benchmarking,henderson2018deep,mania2018simple,chua2018deep}.
Second, the rationale behind \iac{NN} policy's actions is obscured by its complex parameterization, limiting interpretability.
Third, constraint handling is not intrinsic to the \ac{NN} structure.
Instead, constraints are typically encouraged through penalties or satisfied only in expectation \cite{altman1999constrained,brunke2022safe}, so hard guarantees generally require an additional safety layer, such as a control barrier function or a predictive safety filter \cite{ames2019control,wabersich2021predictive,wabersich2023data-driven}.

\textbf{Model-based policies.}\quad Model-based policies offer an alternative that can mitigate these limitations by computing actions using models of the system dynamics, cost, and potentially constraints.
This class spans a wide range of model-based control approaches, including \ac{LQR}, \ac{MPC} \cite{rawlings2017model}, \ac{MPPI} \cite{williams2016aggressive}, and optimization layers such as safety filters \cite{ames2019control,wabersich2021predictive}.
For concreteness, we use \ac{MPC} as a representative model-based policy, owing to its maturity in constrained, multivariable control and its well-established control-theoretic properties, such as stability and recursive feasibility \cite{mayne2000constrained,bertsekas2005dynamic}.

\begin{figure}[t]
    \centering
    \includegraphics[width=0.45\textwidth]{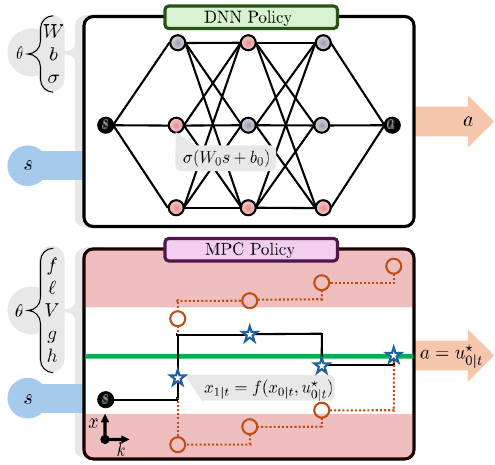} 
    \caption{
    A conceptual representation of \iac{NN} policy and a model-based policy instantiated via \ac{MPC}.
    The \ac{NN} policy comprises multiple hidden layers, defined by weights $W$, biases $b$, and activation function $\sigma$.
    Receiving input state $s$, each layer of neurons transforms its input to ultimately yield action $a$. 
    The depicted \ac{MPC} policy is parameterized by dynamics model $f$, stage cost $\ell$, terminal cost $V$, and constraints $g$ and $h$.
    Using these models, the policy plans a sequence of actions towards achieving its goal state (green line) while respecting constraints (red area).
    The policy outputs action $a=u_{0|t}^{\star}$ from the optimal sequence (solid \& starred), disregarding suboptimal or infeasible sequences (dashed \& circled).
    }
    \label{fig: policy rep}
\end{figure}

Following control convention, we denote predicted states and input actions by $x$ and $u$, respectively.
Inputs are designed to minimize costs, rather than maximize rewards.
In its nominal form, \ac{MPC} involves the solution to the following optimal control problem at each measured state $s$:
\begin{subequations} \label{eqn: mpc problem}
    \begin{alignat}{2}
        \min_{u_{0:H-1|t}} & \sum_{k=0}^{H-1} \ell(x_{k|t},u_{k|t}) + V(x_{H|t}) \label{eqn: MPC objective} \\
        \text{s.t.} \hspace{3pt} & x_{k+1|t} = f(x_{k|t},u_{k|t}) \label{eqn: MPC model} \\
        & g(x_{k|t},u_{k|t}) = 0 \label{eqn: MPC equality constraint}\\
        & h(x_{k|t},u_{k|t}) \leq 0 \label{eqn: MPC inequality constraint}\\
        & x_{0|t} = s \label{eqn: MPC initial state}\\
        & \forall~k = 0,\ldots,H-1, \label{eqn: MPC all time}
    \end{alignat}
\end{subequations}
where the optimal inputs $u_{0:H-1|t}^{\star} = [u_{0|t}^{\star}, \dots, u_{H-1|t}^{\star}]$ minimize the stage costs $\ell: \mathbb{R}^n \times \mathbb{R}^m \to \mathbb{R}$ and terminal cost $V: \mathbb{R}^n \to \mathbb{R}$ over the finite horizon $H$, subject to the dynamic model $f: \mathbb{R}^n \times \mathbb{R}^m \to \mathbb{R}^n$ and the equality and inequality constraints $g: \mathbb{R}^n \times \mathbb{R}^m \to \mathbb{R}^{n_g}$ and $h: \mathbb{R}^n \times \mathbb{R}^m \to \mathbb{R}^{n_h}$.
At each state measurement $s$, \cref{eqn: mpc problem} is solved, but only the first input $u_{0|t}^{\star}$ is applied to the system before repeating the process with the next measurement.
This receding-horizon approach to optimal control defines an \emph{implicit} model-based policy
\begin{equation} \label{eqn: MPC policy}
    \pi_{\theta}^{\text{MPC}}(s) = u_{0|t}^{\star},
\end{equation}
where $\theta$ parameterizes any or all of $\ell, V, f, g,$ and $h$ in \cref{eqn: MPC objective,eqn: MPC model,eqn: MPC equality constraint,eqn: MPC inequality constraint}.
The differences in parameterization and evaluation of \ac{NN} and \ac{MPC} policies are depicted in \Cref{fig: policy rep}.
While \iac{NN} explicitly represents a single global policy for evaluation at any state, \iac{MPC} policy is defined implicitly, solving a trajectory optimization at each state encountered.

The structure of \cref{eqn: mpc problem} brings numerous benefits to policy optimization:
\begin{itemize}
    \item \emph{Interpretability:}\quad Unlike a black-box \ac{NN}, the equation-oriented structure of \cref{eqn: mpc problem} renders the model-based policy's decisions interpretable.
    The objective and constraints are defined explicitly, and each action is accompanied by a predicted trajectory and a set of active constraints.
    \item \emph{Safety:}\quad Actions are designed based on predicted trajectories that obey the model dynamics and satisfy constraints encoding system safety and operating limits. 
    Robust \ac{MPC} formulations extend these properties to uncertainty in the true system, guaranteeing constraint satisfaction, recursive feasibility, and stability under suitable assumptions \cite{mayne2000constrained,bemporad2007robust}.
    \item \emph{Modularity:}\quad Each component of the \ac{MPC} parameterization can be fixed, learned, or replaced
    independently; for example, a physics-based dynamics model can be retained while the cost is learned.
    \Acp{NN}, by contrast, offer no comparably direct way to incorporate prior knowledge of the system.
\end{itemize}

These benefits stem from the inductive bias of a model-based policy representation.
An \ac{NN} makes minimal assumptions about the system, whereas the components of \cref{eqn: mpc problem} can be designed to approximate the system and task at hand: \cref{eqn: MPC model} models the system dynamics, the stage cost $\ell$ models the negative reward, and the terminal cost $V$ approximates the negative value function \cite{mayne2000constrained,lawrence2025view}.
When these models are sufficiently accurate, the policy plans trajectories toward high-reward regions of the state-action space while avoiding unsafe regions encoded by \cref{eqn: MPC equality constraint,eqn: MPC inequality constraint}.
Conversely, \cref{eqn: MPC policy} can still achieve high performance even when its component models are individually inaccurate; for example, given an inaccurate dynamics model, richly parameterized stage and terminal costs can be learned to compensate, such that the resulting \ac{MPC} policy remains optimal \cite{gros2020data-driven}.
Nonetheless, each component of a model-based policy can further be restricted to a model class informed by the physics of the system, or fixed when known with certainty.
Such choices would reduce the number of learnable parameters relative to \iac{NN} policy, confining the search to a lower-dimensional set of physically meaningful parameterizations and thereby reducing the samples required to learn a performant policy.
Consequently, a model-based policy starts from an informed initial guess rather than a random initialization, and learning needs to only correct errors in the component models.
Learning these components for closed-loop performance embodies the notion of \ac{I4C}, in which \cref{eqn: performance function} dictates the identification criterion; the policy is designed to account for model inaccuracies in a performance-oriented manner \cite{gevers1993towards}.

\subsection{Differentiating Model-based Policies} \label{sec: 3.2}

Gradient-based approaches to model-free learning, such as policy gradient \ac{RL}, require the policy to be differentiable with respect to its parameters.
While \iac{NN} policy is an explicit function whose derivatives can be obtained from automatic differentiation, the model-based policy \cref{eqn: MPC policy} is defined implicitly as the solution of an optimization problem, which makes differentiating the policy a central challenge. 
One approach to computing the derivatives of \cref{eqn: MPC policy} is to unroll the iterations of the numerical solver and apply automatic differentiation to the resulting computational graph \cite{ji2021bilevel}.
However, this process is computationally and memory intensive, and may suffer from vanishing or exploding gradients as the number of solver iterations increases \cite{pineda2022theseus}.
These limitations arise because unrolling computes gradients by differentiating through the iterative solution procedure, rather than directly differentiating the optimality conditions that characterize the solution.
The \acf{IFT} \cite{dontchev2009implicit} addresses this challenge by providing a way to differentiate the optimality conditions at the solution, independently of the numerical procedure used to obtain it. 
Under suitable regularity conditions, including strict complementarity and the nonsingularity of the Jacobian of the optimality conditions with respect to the primal and dual variables, the \ac{IFT} guarantees that the solution is locally a unique, differentiable function of the learnable parameters $\theta$, and provides an expression for its derivative.

\begin{theorem}[\Acl{IFT} \cite{dontchev2009implicit}]
Let $\psi: \mathbb{R}^{d} \times \mathbb{R}^{p} \rightarrow \mathbb{R}^{d}$ be continuously differentiable in the unknown $\xi \in \mathbb{R}^d$ and parameter $\vartheta \in \mathbb{R}^p$, and let $(\bar{\xi}, \bar{\vartheta})$ satisfy $\psi(\bar{\xi}, \bar{\vartheta}) = 0$ with $\nabla_{\xi}\psi(\bar{\xi}, \bar{\vartheta})$ nonsingular.
Then, there exist neighborhoods $\mathcal{U}$ of $\bar{\vartheta}$ and $\mathcal{W}$ of $\bar{\xi}$ such that, for each $\vartheta \in \mathcal{U}$, there is a unique $\xi^{\star}(\vartheta) \in \mathcal{W}$ satisfying $\psi(\xi^{\star}(\vartheta), \vartheta) = 0$.
Moreover, $\xi^{\star}$ is continuously differentiable on $\mathcal{U}$, with $\xi^{\star}(\bar{\vartheta}) = \bar{\xi}$ and Jacobian
\begin{equation} \label{eqn: IFT}
    \nabla_{\vartheta}\xi^{\star}(\vartheta) = -\big[\nabla_{\xi}\psi(\xi^{\star}(\vartheta), \vartheta)\big]^{-1} \nabla_{\vartheta}\psi(\xi^{\star}(\vartheta), \vartheta), \quad \forall\, \vartheta \in \mathcal{U}.
\end{equation}
\end{theorem}
Notably, the \ac{IFT} does not require tracking intermediate iterates for derivative calculations, reducing the overall memory usage \cite{pineda2022theseus}.
To apply the \ac{IFT} to \cref{eqn: MPC policy}, we first define the Lagrangian of \cref{eqn: mpc problem}:
\begin{equation*}\begin{aligned}
    \mathcal{L}(z,\theta;s) = \sum_{k=0}^{H-1} \Big( & \ell(x_{k|t},u_{k|t}) + \nu_{k}^{\top} g(x_{k|t},u_{k|t}) \\
    & + \lambda_{k}^{\top} h(x_{k|t},u_{k|t}) \Big) + V(x_{H|t}),
\end{aligned}\end{equation*}
where the states $x_{k|t}$ are eliminated by recursively substituting \cref{eqn: MPC model}; $z = (u_{0:H-1|t}, \nu, \lambda)$ contains both the primal and dual variables, i.e., the Lagrange multipliers $\nu = (\nu_{0}, \ldots, \nu_{H-1})$ and $\lambda = (\lambda_{0}, \ldots, \lambda_{H-1})$; and $\theta$ enters through any of $\ell$, $V$, $f$, $g$, and $h$.
The optimal solution $z^{\star} = (u_{0:H-1|t}^{\star}, \nu^{\star}, \lambda^{\star})$ satisfies the \ac{KKT} conditions of \cref{eqn: mpc problem}
\[
    \psi(z^{\star},\theta;s) = \begin{bmatrix}
        \nabla_{u_{0:H-1|t}}\mathcal{L}(z^\star,\theta;s) \\
        g(x_{0|t},u_{0|t}^{\star}) \\
        \vdots \\
        g(x_{H-1|t},u_{H-1|t}^{\star}) \\
        \text{diag}(\lambda_{0}^{\star})~h(x_{0|t},u_{0|t}^{\star}) \\
        \vdots \\
        \text{diag}(\lambda_{H-1}^{\star})~h(x_{H-1|t},u_{H-1|t}^{\star}) \\
    \end{bmatrix} = 0,
\]
where the conditions $\lambda_k^\star \ge 0$ and $h(\cdot) \le 0$ are omitted since, under strict complementarity, the set of active constraints does not change for small perturbations of $\theta$.

The \ac{KKT} conditions implicitly define $z^\star$ as a function of $\theta$, which can be differentiated to acquire sensitivities of $z^{\star}$ via \cref{eqn: IFT} by setting $\xi = z \in \mathbb{R}^{d}$ and $\vartheta = \theta \in \mathbb{R}^{p}$ for a given state $s$.
Solving the resulting linear system yields the sensitivities, from which the derivatives of the \ac{MPC} policy are extracted as the rows corresponding to $u_{0|t}$:
\begin{equation} \label{eqn: MPC sensitivity}
    \nabla_{\theta}\pi_{\theta}^{\text{MPC}}(s) = -\big[[\nabla_{z}\psi(z^{\star},\theta;s)]^{-1} \nabla_{\theta}\psi(z^{\star},\theta;s)\big]_{u_{0|t}} ~ ,
\end{equation}
provided that $z^{\star}$ satisfies \acl{LICQ}, \acl{SOSC}, and strict complementarity, which together ensure that $\nabla_z \psi(z^\star, \theta; s)$ is nonsingular \cite{fiacco1976sensitivity}.
However, the resulting derivatives $\nabla_{\theta}\pi_{\theta}^{\text{MPC}}(s)$ have three limitations: they may not exist where the active set changes; they are approximate when the \ac{MPC} problem \cref{eqn: mpc problem} is solved inexactly; and they are local to the solution returned by the solver when the problem is nonconvex.
Smoothing or regularization techniques \cite{zuliani2025differentiable} and unrolling or error-aware implicit differentiation \cite{ablin2020super,blondel2022efficient} can mitigate some of these limitations, but addressing all three simultaneously remains an open challenge.
Computing the sensitivities can also be expensive, motivating methods that accelerate or circumvent this computation \cite{amos2018differentiable,frey2025differentiable,lawrence2025mpcritic}.

Despite these challenges, the \ac{IFT} exploits the structure of the \ac{MPC} problem to quantify how the learnable parameters $\theta$ of a model-based policy influence decisions; information that \ac{DFO} methods discussed below, which treat the policy as a black box, do not use.
Although presented here for nominal \ac{MPC}, similar techniques based on the \ac{IFT} extend to robust \ac{MPC} \cite{kordabad2021reinforcement}, mixed-integer \ac{MPC} \cite{gros2020reinforcement}, and moving horizon estimators \cite{esfahani2022policy}.
These sensitivities form the basis of the gradient-based methods for model-based policies discussed next.

\subsection{Policy Optimization for Model-based Policies} \label{sec: 3.3}

\begin{figure*}[t!]
    \centering
    \includegraphics[width=\textwidth]{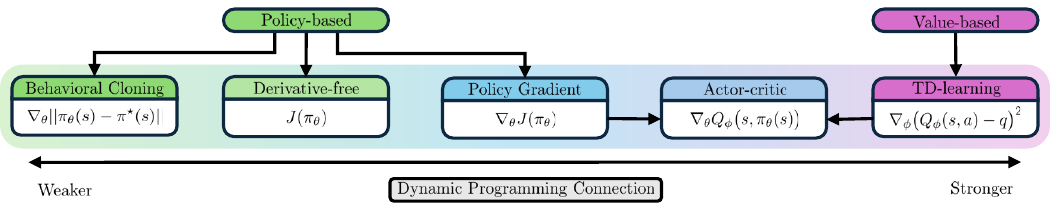}
    \caption{
    The spectrum of approaches to designing policies and their connection to \ac{DP}.
    Policy-based approaches directly optimize a policy function approximator $\pi_\theta$ and need not employ ideas from \ac{DP}.
    Furthest removed, behavioral cloning makes no use of value, instead mimicking an expert's actions, e.g., those of an optimal policy $\pi^\star(s)$.
    \Acl{DFO} targets the performance function $J(\pi_\theta)$ but assumes little to no knowledge of the system or policy structure.
    Policy gradient methods exploit the \ac{MDP} structure to estimate $\nabla_\theta J(\pi_\theta)$.
    Actor-critic methods, including most offline \ac{RL} methods, estimate these gradients with the help of value function approximators learned via \ac{TD} learning, which is grounded in the Bellman equation.
    At the far end, value-based methods such as $Q$-learning use \ac{TD} learning to learn the optimal $Q$-function itself, with the policy defined in terms of the learned $Q$-function.
    }
    \label{fig: policy optimization}
\end{figure*}

We now turn to the central question of this paper: how the parameters $\theta$ of a model-based policy can be learned in a model-free manner.
Because the relationship between the components of a model-based policy and the resulting closed-loop behavior generally cannot be described analytically, $\theta$ must be adapted from data.
The information the data provide determines which methods apply and how closely each relates to \ac{DP}, as depicted in \Cref{fig: policy optimization}.
We distinguish two classes of methods:

\begin{enumerate}[(i)]
\item \emph{derivative-free} methods, which only require noisy evaluations of the performance function $J(\pi_\theta)$ and are, therefore, agnostic to whether the policy is defined explicitly or implicitly; and
\item \emph{gradient-based} methods, which require derivatives of the policy with respect to $\theta$.
These derivatives may not be trivial to obtain for implicitly defined policies such as \cref{eqn: MPC policy}, as discussed above.
\end{enumerate}
Gradient-based methods are further distinguished by the data they learn from:
\begin{itemize}
\item \emph{policy gradient} methods estimate the gradient of the performance function, $\nabla_\theta J(\pi_\theta)$, from online system interaction via the deterministic or stochastic \ac{PGT}, using either complete trajectories and their returns or individual reward-labeled transitions aided by a learned value function;
\item \emph{offline \ac{RL}} methods learn from a fixed dataset of previously collected reward-labeled transitions, typically with actor-critic techniques regularized toward the behavior in the dataset; and
\item \emph{\acl{BC}} methods learn from a fixed dataset of expert demonstrations by minimizing a supervised loss on expert state-action pairs, requiring neither system interaction nor rewards.
\end{itemize}
We briefly outline each class below, with detailed treatments following in \Cref{sec: 4,sec: 5,sec: 6}.

\textbf{Derivative-free methods.}\quad When gradient information about the performance function cannot be practically obtained, \ac{DFO} solves the policy search problem \cref{eqn: policy search} directly from evaluations of $J(\pi_{\theta})$.
Because \ac{DFO} is agnostic to the differentiability of the policy and the performance function, it is a natural choice for implicitly defined policies such as the \ac{MPC} policy \cref{eqn: MPC policy}.
This black-box treatment, however, connects \ac{DFO} only indirectly to \ac{DP}, and \ac{DFO} is typically restricted to episodic settings in which performance is evaluated over a finite horizon.
Moreover, common \ac{DFO} methods, such as random search \cite{solis1981minimization}, genetic algorithms \cite{katoch2021review}, and CMA-ES \cite{hansen2003reducing}, can be sample-inefficient.
This inefficiency matters little when policy evaluations are cheap and parallelizable, but becomes prohibitive when each evaluation of $J(\pi_{\theta})$ requires a physical experiment or a high-fidelity simulation \cite{shahriari2015taking}. 

\ac{BO} \cite{frazier2018tutorial} addresses this sample inefficiency.
\Ac{BO} is a global optimization method for expensive, noisy, black-box functions: it fits a probabilistic surrogate of the performance function from observed evaluations and selects each new query to balance exploration of the decision space against exploitation of promising regions, seeking a global optimum within a limited evaluation budget.
Because the surrogate makes minimal assumptions about the performance function and the underlying closed-loop dynamics, \ac{BO} applies directly to a wide range of model-based policy formulations.
\Ac{BO} also readily extends beyond standard \ac{MDP} formulations, accommodating multiple performance objectives, time-varying conditions, and sparse rewards \cite{makrygiorgos2022performance-oriented,shao2024time-varying,shao2024coactive}, as well as constraints on closed-loop behavior.
These properties make \ac{BO} particularly well suited to learning model-based policies with a moderate number of learnable parameters \cite{paulson2023tutorial}.
    
\textbf{Gradient-based methods.}\quad When the policy is differentiable with respect to $\theta$, the gradient of the performance function $\nabla_\theta J$ can be estimated from sampled trajectories, or replaced by the gradient of a surrogate objective such as a supervised loss on expert state-action pairs.
Estimates of $\nabla_\theta J$ are possible because the performance function is an expected cumulative reward, which allows the \ac{PGT} to express its gradient in terms of derivatives of the policy.
For model-based policies, derivatives of the policy as required by the \ac{PGT} are typically established via the \ac{IFT}, which provides the sensitivities in \cref{eqn: MPC sensitivity}.

\emph{Policy gradients.}\quad The main obstacle to computing $\nabla_\theta J(\pi_\theta)$ in solving \cref{eqn: policy search} is that $\theta$ influences not only the actions taken but also the distribution of closed-loop states visited under $\pi_\theta$.
This dependence is unknown since it is governed by the unknown system dynamics. 
By exploiting the Markovian nature of the dynamics, the \ac{PGT} provides an expression for $\nabla_{\theta}J(\pi_{\theta})$, which requires derivatives of the policy but not the state distribution \cite{sutton1999policy}.
Although the expectation in \cref{eqn: performance function} generally prevents computing the policy gradient exactly, the gradient can be estimated from system samples for stochastic gradient ascent on $J(\pi_\theta)$, using either Monte Carlo returns or learned value functions in actor-critic methods.

Monte Carlo estimation of the policy gradient requires only the ability to sample closed-loop trajectories, evaluate their returns, and differentiate the policy via \cref{eqn: MPC sensitivity} for model-based policies.
Like \ac{DFO}, Monte Carlo policy gradient methods require neither the Bellman equation nor value functions, and are episodic in nature.
However, methods such as REINFORCE \cite{williams1992simple} are generally restricted to stochastic policies, so the deterministic MPC policy \cref{eqn: MPC policy} must first be made stochastic, for example, by adding Gaussian exploration noise to its output.
Furthermore, these methods can suffer from high variance and slow learning \cite{sutton2018reinforcement}.

To reduce this variance, actor-critic methods learn both a policy $\pi_{\theta}$, the \emph{actor}, and a separately parameterized value function $Q_{\phi}$, the \emph{critic} \cite{grondman2012survey}.
The critic is learned through the Bellman equation to predict the infinite-horizon value of states and actions, giving actor-critic methods a closer connection to \ac{DP} than Monte Carlo policy gradients.
The actor is then updated to maximize these predictions via the stochastic or deterministic \ac{PGT} \cite{sutton1999policy,silver2014deterministic}.
Hence, actor-critic methods can admit deterministic policies such as \cref{eqn: MPC policy} \cite{silver2014deterministic}.
Because the critic is learned from individual state transitions rather than entire trajectories, actor-critic methods can accommodate continuing tasks without terminating trajectories \cite{grondman2012survey}.
They can also reuse samples from previous policies and substantially reduce the variance of gradient estimates \cite{lillicrap2016continuous,schulman2018high-dimensional}.
However, learning the critic can introduce instability due to value function approximation, bootstrapping, and off-policy learning \cite{sutton2018reinforcement}.

\emph{Offline \ac{RL}.}\quad The preceding methods require online interaction with the system, which may be expensive or unsafe.
In many applications, historical data are already available and could inform the policy without further interaction.
Offline \ac{RL} learns policies from such a fixed dataset of previously collected transitions \cite{levine2020offline}.
Offline \ac{RL} builds on the same approximations as off-policy actor-critic methods, but must account for distribution shift: the learned policy may select actions absent from the dataset, where the critic's estimates are unreliable and tend to be overestimated.
Methods based on conservative value estimation \cite{kumar2020conservative}, regularization toward the dataset's behavior \cite{fujimoto2021minimalist}, and in-sample learning \cite{kostrikov2022offline} mitigate this issue.
Even so, the attainable performance depends on how well the dataset covers the relevant states and actions; thus, diverse, well-excited data are needed \cite{schweighofer2022dataset}.
Compared with online actor-critic methods, the combination of offline \ac{RL} with model-based policies remains relatively unexplored \cite{sawant2023learning-based}.

\emph{Behavioral cloning.}\quad The preceding methods assume a well-specified performance function $J(\pi_\theta)$, which can be difficult to formulate, for example, when the goal is to reproduce the behavior of a human operator.
Inverse \ac{RL} first infers a reward function that explains expert demonstrations and then optimizes a policy to maximize it \cite{osa2018algorithmic}.
However, the inference problem is ill-posed, since many rewards explain the same behavior \cite{ng2000algorithms}, and expensive, since it nests \ac{RL} within reward inference.
\ac{BC} avoids reward inference by regressing the expert's actions on the states visited in the demonstrations.
By reducing policy learning to supervised learning, \ac{BC} provides a pragmatic way to design expert-like policies, requiring only an expert dataset and derivatives of the policy.
However, the learned policy is at best as good as the demonstrations.
Any mismatch with the expert can drive the system into states absent from the dataset, where errors compound \cite{ross2011reduction}.
Nonetheless, owing to its simplicity, \ac{BC} was among the first settings in which model-based policies were learned with gradient-based methods \cite{amos2018differentiable}.
The structure of model-based policies, including constraints and predictive models, may help contain the compounding errors that limit \ac{NN} policies in \ac{BC} \cite{east2020infinite-horizon}.
        
\section{Bayesian Optimization} \label{sec: 4}

This section presents \ac{BO} as a derivative-free approach to solving the policy optimization problem in \cref{eqn: policy search}.
We first outline the core algorithmic components of \ac{BO} for single-objective optimization.
We then discuss the features that make \ac{BO} well suited to model-free learning of model-based policies, followed by highlighting open research opportunities in this area.

\subsection{Algorithmic Components of \ac{BO}}
\begin{figure}[htb]
\centering
\includegraphics[width=0.9\linewidth]{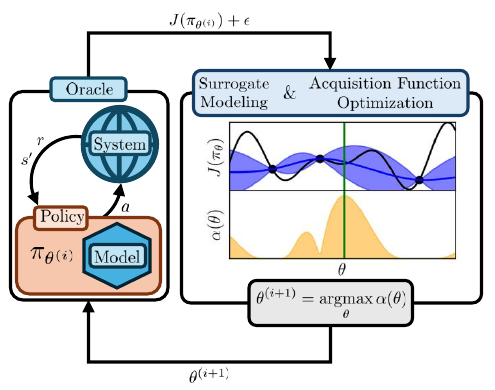}
\caption{
Policy optimization via \ac{BO} is episodic in nature, relying on zeroth-order information of objective $J(\pi_{\theta})$.
The oracle, composed of the parametric policy $\pi_{\theta^{(i)}}$ and the system, is treated as a black-box function.
With each episode $i$, the oracle produces an observation of $J(\pi_{\theta})$ corrupted by random noise $\epsilon$.
These observations are used to construct a probabilistic surrogate model (blue) of unknown true $J(\pi_{\theta})$ (black).
The surrogate model is used to define the \ac{AF} $\alpha(\theta)$ (orange), which is maximized to choose the next policy parameters $\theta^{(i+1)}$ (green).
The oracle is queried at the chosen $\theta$, and the process repeats until the budget for learning is exhausted.
}
\label{fig: Bayesian optimization}
\end{figure}

As a \ac{DFO} method, \ac{BO} aims to solve the global optimization of black-box functions, which may be expensive to evaluate, through a sequence of zeroth-order queries of the closed-loop system \cite{frazier2018tutorial}.
Taking a Bayesian view, \ac{BO} uses the full history of function observations when deciding where to query next.
\Ac{BO} iteratively re-queries the function evaluation oracle, i.e., the closed-loop system composed of the learned policy and system, until a pre-defined stopping rule is met.
A visual representation of this process can be seen in \Cref{fig: Bayesian optimization}.

Critically, \ac{BO} considers the performance function $J(\pi_{\theta})$ to be continuous with minimal assumptions regarding its structure or functional form, encoded through the prior \cite{frazier2018tutorial}.
The Bayesian approach is to model the objective as a random function based upon a prior capturing beliefs of the function's behavior.
Encompassed by a \emph{probabilistic surrogate model}, newly obtained observations of the closed-loop performance from re-querying the function evaluation oracle can update this prior, thereby informing the posterior distribution over the performance function.
Using the modeled posterior distribution, an \emph{\acl{AF}} (\ac{AF}) is constructed for deciding where to next query the oracle.
An \ac{AF} is described by a user-selected objective that quantifies the utility of a decision towards maximizing the underlying black-box function.

\textbf{Probabilistic surrogate model.}\quad A critical feature of a probabilistic surrogate model for \ac{BO} is uncertainty quantification.
This feature enables modeling the posterior distribution of function values $J(\pi_{\theta})$ at decisions $\theta$.
To this end, \acp{GP} \cite{frazier2018tutorial} are the most commonly used surrogate models in \ac{BO}, but Bayesian neural networks \cite{li2024study}, deep \acp{GP} \cite{hebbal2023deep}, random forests \cite{hutter2011sequential}, and dropout \acp{NN} \cite{guo2022evolutionary} have all been proposed as alternative surrogate model classes.
A distinguishing feature of \acp{GP} is their nonparametric statistical nature, placing minimal assumptions on the black-box function, i.e., $J$, from which data are obtained \cite{rasmussen2008gaussian}.
As opposed to assuming the function belongs to a finitely parametrized space, \acp{GP} describe a probability distribution whose support is over the space of continuous functions \cite{rasmussen2008gaussian}.

\textbf{Acquisition function.}\quad The \ac{AF} generally involves a trade-off between performance improvement in optimizing $J(\pi_{\theta})$ and uncertainty reduction about $J(\pi_{\theta})$ \cite{frazier2018tutorial}.
It is through decisions that maximize the \ac{AF} that \ac{BO} aims to strike a balance between the competing objectives: exploration (i.e., querying where posterior predictions are most uncertain) and exploitation (i.e., querying where posterior predictions are confident and high-performing).
The decision-making process for determining the next query can be described by the optimization of an \ac{AF} $\alpha(\theta)$:
\[
\theta^{(i+1)} = \argmax_{\theta \in \Theta} \alpha(\theta).
\]
Common \acp{AF} include (log) expected improvement, probability of improvement, upper confidence bound, knowledge gradient, and entropy search \cite{frazier2018tutorial,daulton2023unexpected}.
Moreover, a variety of \acp{AF} exist for problems beyond the single-objective black-box optimization, as discussed in the ensuing. 

As applied to the policy optimization problem, \ac{BO} treats $J(\pi_{\theta})$ as a black-box function, ignoring much of the underlying \ac{MDP} problem structure discussed in \Cref{sec: 2}.
As a consequence, the mathematical structure of \cref{eqn: performance function}, stemming from the sequential decision-making process, is not exploited in standard \ac{BO}.
Rather, \ac{BO} makes use of noisy observations from the closed-loop system, i.e., $y = J(\pi_{\theta}) + \epsilon$, where $\epsilon$ is the observation noise for the performance function when approximating the expectation of \cref{eqn: performance function} due to the stochastic policy, dynamics, and initial state.
Although \ac{BO} does not necessarily exploit the \ac{MDP} problem structure, it can utilize prior observations $\mathcal{D}^{N} = \{(\theta^{(i)}, y^{(i)})\}_{i=0}^{N-1}$ in constructing the posterior to avoid starting control policy optimization from scratch.
However, improving the policy requires interaction via iteratively querying the function evaluation oracle according to the chosen \ac{AF} and updating the posterior. 

\subsection{Strengths for Policy Optimization}
Two strengths make \ac{BO} particularly well suited to policy optimization: its inherent ability to \emph{balance trade-offs} between competing objectives, and its \emph{derivative-free flexibility}.

\textbf{Balancing trade-offs.}\quad \acp{AF} are designed to balance competing objectives in a principled way, most notably exploration and exploitation.
The same principle extends naturally to policy optimization problems with competing goals.
These include \emph{\ac{MO} optimization}, which involves multiple performance functions $\{J_{j}\}_{j=0}^{N_{J}-1}$, and \emph{constrained optimization}, where maximizing performance may conflict with satisfying constraints.

\emph{\Acl{MO} optimization:}\quad Designing an optimal control policy for multiple objectives typically does not have a single optimal solution.
Rather, \ac{MO} optimization problems often seek to map the Pareto front: the set of Pareto optimal solutions such that any improvement in one objective deteriorates another \cite{miettinen1999nonlinear}.
Having knowledge of the Pareto front enables informed policy design, spanning potentially multiple trade-offs.
Evolutionary algorithms are a common approach to \ac{MO} problems \cite{deb2002fast}, but \ac{BO} provides a sample-efficient alternative for balancing \ac{MO} optimization \cite{daulton2020differentiable}.
A common approach is to formulate \iac{MO}-\ac{AF} defined by a hypervolume indicator \cite{guerreiro2021hypervolume}.
The hypervolume indicator measures the volume of the objective space dominated by the observed Pareto front relative to a reference point, quantifying the quality of the Pareto front.

With a growing number of sophisticated \ac{MO}-\ac{BO} algorithms and \acp{AF} \cite{daulton2020differentiable,daulton2022multi-objective,daulton2023unexpected}, their application to policy optimization with multiple objectives can be seen in a variety of settings.
In \cite{makrygiorgos2022performance-oriented}, \ac{MO}-\ac{BO} is used to learn a performance-oriented \ac{MPC} model, balancing two conflicting control objectives for a bioreactor system.
This is accomplished by iteratively suggesting and trialing policies expected to be Pareto dominating, which would grow the hypervolume indicator \cite{daulton2020differentiable}.
Similarly, \cite{hoang2025multi-objective} learn the \ac{MPC} objective to balance wind farm equipment loading and reference tracking objectives, using an automatic relevance determination algorithm to effectively prune the parameter search space and simplify the policy optimization process.
An \ac{MO}-\ac{BO} formulation is also demonstrated for encapsulating trade-offs in performance, safety, and hardware implementation objectives \cite{sorourifar2021data,chan2024practical}, towards learning an approximate \ac{MPC} policy for medical application as an example.
These works exemplify the range of policy optimization problem formulations and system descriptions that \ac{MO}-\ac{BO} can accommodate for model-based policies.

\emph{Constrained optimization:}\quad While the goal of policy optimization is to maximize the total reward, safety-critical systems, such as those in medical and robotics applications \cite{brunke2022safe,chan2023safe}, require that the policy respects safety constraints.
This becomes a challenge when maximizing \cref{eqn: performance function} is in conflict with the policy satisfying system constraints, which themselves can be treated as black-box functions. Constrained \ac{BO} offers multiple approaches to the constrained policy optimization problem, depending on the required assurances of constraint satisfaction.
Early constrained \ac{BO} methods learn a probabilistic surrogate model for constraints and encode the probability of constraint violation into the acquisition function \cite{schonlau1998global,gardner2014bayesian}.
\cite{sorourifar2021data} and \cite{gharib2021multi-objective} demonstrate how this approach can be applied to learn \ac{MPC} policy parameters for constrained single and multi-objective problems. 
However, there are no guarantees on constraint satisfaction for this class of methods \cite{xu2023constrained}.

There have been significant developments in constrained \ac{BO} in recent years aimed at providing constraint satisfaction guarantees, which can be broadly classified into penalty-based and safe set strategies.
Penalty-based strategies augment an \ac{AF} with penalty terms on constraint violation \cite{lu2022no-regret}.
This strategy has been applied to learning \ac{MPC} policies with closed-loop stability requirements \cite{hirt2024stability-informed}, but penalty-based strategies are generally sensitive to the choice of the penalty weight and can become stuck locally as a consequence.
On the other hand, safe set strategies rely on the notion of establishing an estimated safe region for the policy parameters, constraining the \ac{AF} to this region using tools such as Lipschitz continuity  \cite{berkenkamp2023bayesian} or log barrier functions \cite{krishnamoorthy2023model-free}.
To ensure constraint satisfaction with high probability, these strategies require a well-calibrated surrogate of constraints in estimating the safe region.
However, balancing exploration and exploitation remains a challenge in constrained \ac{BO} based on safe sets.
To this end, \cite{chan2023safe} propose a safety-based \ac{AF} that enhances safe exploration by enlarging the safe region.

\textbf{Derivative-free flexibility.}\quad Treating the performance function as a black box gives \ac{BO} considerable flexibility.
Using only performance observations, \ac{BO} can optimize any continuous function whose regularity is reasonably captured by the surrogate's prior; the mapping from policy parameters to the objective need not be available in closed form, have accessible derivatives, or exhibit any other known structure.
As a result, \ac{BO} enables \emph{policy-agnostic optimization}, placing few restrictions on how the policy is defined, and extends \emph{beyond \acp{MDP}} to reward and dynamics definitions outside the standard formulation of \Cref{sec: 2}.

\emph{Policy-agnostic optimization:}\quad Because \ac{BO} does not exploit structural knowledge of the relationship between policy parameters and performance function, it can learn any general control policy representation.
This includes policies defined over discrete or mixed parameter spaces \cite{daxberger2020mixed-variable,daulton2022bayesian}, which are not naturally accommodated in gradient-based optimization.
Furthermore, \ac{BO} is particularly useful for learning policies that are not readily differentiable, or whose gradient estimates are unreliable.

Exemplifying these ideas for \ac{MPC} policies, \ac{BO} does not require that learnable parameters explicitly appear in an optimization-based policy, or in its \ac{KKT} conditions.
The dynamics model, constraints, and costs can be treated just the same as the prediction or robust horizon, \ac{MPC} optimizer settings, or other \ac{MPC} design choices that do not explicitly appear in the control law \cref{eqn: MPC policy}.
This differs from gradient-based learning algorithms that exploit the \ac{KKT} conditions to calculate gradients, as described in \Cref{sec: 3.2}.
As an example, \cite{piga2019performance-oriented} learn the parameters of a hierarchical architecture in which an \ac{MPC} policy serves as the reference governor for a PID controller.
In this scheme, \ac{BO} learns the dynamics model of an \ac{MPC} policy, appearing in \ac{KKT} conditions, and its control horizon and the gains of a lower-level PID controller, which do not.
While complex policy representations such as mixed-integer nonlinear programming require particular care to learn with gradient-based strategies \cite{gros2020reinforcement}, \ac{BO} can be applied directly out of the box to such representations \cite{choksi2021simulation-based} without concern for active-set changes or violations of \acl{LICQ}, \acl{SOSC}, or strict complementarity, which complicate gradient-based learning.

\emph{Beyond \acp{MDP}:} The standard \ac{MDP} framework described in \Cref{sec: 2} does not encompass every possible control policy optimization scenario.
The standard \ac{MDP} assumptions regarding the dynamics, reward function, or the overall observability of the process may not hold in a general policy optimization problem.
However, due to its flexible nature, \ac{BO} remains largely applicable to such scenarios.
Two relevant examples are control policy optimization in time-varying systems \cite{bogunovic2016time-varying} and from preferential feedback as opposed to a well-defined reward signal \cite{akrour2011preference-based}. Many real-life systems can exhibit gradual drifts in behavior, for example, due to aging components, or persistent time-varying environmental disturbances.
Time-varying \ac{BO} can remedy this issue by learning both the spatial and temporal correlations between queries and their outcomes \cite{bogunovic2016time-varying}.
As such, time-varying \ac{BO} can learn policies for systems with time-varying parameters as in \cite{brunzema2022on}, and those that vary between discrete episodes like that of \cite{shao2024coactive,cho2024run-indexed}, by learning nonstationary surrogate models.

Another practical issue with policy optimization in a standard \ac{MDP} framework arises from the need to specify a reward function.
Specifying a reward function that prevents ``reward hacking'' \cite{amodei2016concrete}, is well ``shaped'' to efficiently guide policy optimization such as a potential-based reward \cite{ng1999policy}, and effectively balances trade-offs in the performance objective(s) or safety requires considerable system expertise.
Policy optimization with an ill-specified reward can result in unintended and unsafe closed-loop control behavior \cite{pan2022effects}.
To this end, preferential \ac{BO} brings the ``human-in-the-loop'' by enabling user interaction to guide policy optimization \cite{gonzalez2017preferential}.
For example, pairwise comparison of policy behavior can constitute useful feedback for policy optimization tasks.
\cite{zhu2021preference-based} apply preferential \ac{BO} for learning model-based policies with examples in chemical processing and autonomous vehicle driving.
\cite{shao2024coactive} expand user-input options in control policy optimization by enabling selection of preferred outcomes and suggestions for alternative outcomes when learning the relationship between the multiple performance functions and the overall utility function.

\subsection{Opportunities for Future Work} \label{sec: 4.3}

While the black-box nature of \ac{BO} offers great flexibility, its minimal structural assumptions on the performance function $J(\pi_{\theta})$ can hinder scalability.
Two important directions for future research are \emph{scalable surrogate modeling} for high-dimensional policy parameterizations, and moving \emph{beyond zeroth-order information} by leveraging trajectory observations, gray-box modeling, or recasting \ac{BO} for policy optimization as an \ac{MDP}.

\textbf{Scalable surrogate modeling.}\quad
\Acp{GP} are the most commonly used surrogate models in \ac{BO}, but diminish in effectiveness as the number of observations of the performance function and learnable policy parameters increases.
Underpinning \acp{GP} are kernel functions that model the covariance between performance values at different policy parameter values, typically as a function of distance between the parameters.
However, the distribution of distances between parameters tightens in higher dimensions as the volume of the search space shifts closer to the boundary, reducing the significance of these distances and deteriorating covariance estimates \cite{binois2022survey}.
This poses a challenge for standard \ac{BO} when a control policy has $\gg 10$ learnable parameters.
Furthermore, the number of observations required to satisfactorily cover the search space increases exponentially with the number of policy parameters.
This is particularly problematic for \acp{GP} as the complexity of inference grows cubically with the number of observations.

Developing methods for handling high-dimensional decision variables is an active area of \ac{BO} research.
\ac{NN}-based surrogate models have shown promise in \ac{BO} due to their efficient inference.
The use of deep kernels allows for combining the nonparametric flexibility of kernel-based methods with the structural properties of \acp{NN} to build surrogate models that scale linearly with the number of observations \cite{wilson2016deep}.
\Acp{BNN} have also shown promise for scalable \ac{BO} \cite{makrygiorgos2025scalable}.
A \ac{BNN} with appropriate priors over network parameters gives rise to a \ac{GP} prior over functions in the infinite limit of hidden units \cite{neal1996priors}.
The resulting \ac{GP} is generally referred to as a neural network \acl{GP} \cite{lee2018deep}, which has several potential benefits over standard \acp{GP}, such as handling nonstationarity and improved performance in high-dimensional parameter spaces \cite{li2024study}.
While the application of neural network surrogates in \ac{BO} for control policy optimization is largely unexplored, these surrogates hold promise for scalable policy optimization while still benefiting from \ac{BO}'s flexibility and ability for balancing trade-offs.

Alternative methods for solving high-dimensional policy optimization problems using \ac{BO} can be broadly classified into the following three categories.
First are methods that assume $J(\pi_{\theta})$ varies along a low-dimensional subspace of the learnable parameters \cite{eriksson2021high-dimensional}, or along low-dimensional embeddings \cite{garnett2013active,wang2016bayesian,lu2018structure,nayebi2019framework}.
These methods are shown to be effective for identifying a subset of \ac{MPC} parameters critical to performance \cite{kudva2024efficient}, or determining linear embeddings that reduce the effective dimensionality of the policy optimization problem \cite{frohlich2019bayesian}.
Second are methods that assume the structure of $J(\pi_{\theta})$, such as being a sum of low-dimensional components \cite{duvenaud2011additive,mutny2018efficient}. 
Third are \ac{BO} methods that follow a local strategy, such as distributing samples across a collection of local regions \cite{eriksson2019scalable}, or incorporating gradient-based parameter updates \cite{muller2021local}.
As an example, \cite{frohlich2021cautious} constrain the \ac{AF} to a confidence region where the surrogate model's predicted uncertainty is low, cautiously exploring  high-dimensional search spaces to locally improve on the initial policy.

\textbf{Beyond zeroth-order information.}\quad
While the minimal assumptions of \ac{BO} about the structure of the policy optimization problem make it a flexible approach, its sample efficiency can be enhanced by leveraging information beyond zeroth-order observations of $J$.
To this end, we consider three approaches: those that leverage closed-loop trajectories, such as for estimating $\nabla_{\theta}J(\pi_{\theta})$; incorporate prior knowledge of the system, i.e., gray-box \ac{BO}; and recast the \ac{BO} problem for policy optimization as \iac{MDP} and subsequently exploit this structure.

In policy optimization, each query can provide observations not only of $J$ but also individual $(s,a,r,s')$ tuples, which contain useful yet rarely utilized information in \ac{BO}.
For example, \cite{wilson2014using} utilize trajectory data to measure the similarity between policies through their behavior rather than parameters, and separately construct a \ac{GP} mean function from learned dynamics and reward models to approximate \cref{eqn: performance function}.
In a similar vein, \cite{kallel2024augmented} construct the \ac{GP} mean function using a learned $Q$-function, incorporating information from past trajectories into the posterior belief about untested policies.
Alternatively, $(s,a,r,s')$ tuples can be used to estimate gradients of \cref{eqn: performance function}, as detailed in \Cref{sec: 5}, useful for both surrogate modeling and \ac{AF} design to accelerate convergence.
\cite{prabuchandran2021novel} develop surrogate models for the gradient and devise a corresponding first-order \ac{AF}.
\cite{makrygiorgos2023no-regret} also devise a single \ac{AF} objective that utilizes zeroth- and first-order information of $J$ to mimic the necessary optimality conditions, while \cite{makrygiorgos2023gradient-enhanced} devise an ensemble of zeroth- and first-order \acp{AF}, selecting parameters via \iac{MO} optimization.

The sample efficiency of \ac{BO} can be enhanced when partial knowledge of the system is available.
Gray-box \ac{BO} leverages knowledge or access to the internal structure of the objective function or constraints \cite{astudillo2021thinking}.
Examples include those that represent the performance function as a composite function and leverage its known parts \cite{lu2022no-regret,lu2023no-regret}, or use low-fidelity approximations of the true system that are cheaper to evaluate to identify potentially promising regions of the parameter space \cite{sorourifar2023computationally,lu2021bayesian}.
\cite{paulson2022cobalt} devise a gray-box \ac{BO} algorithm that exploits the composite structure of the objective and constraints, later extended with no-regret guarantees \cite{lu2022no-regret,lu2023no-regret}.
\cite{sorourifar2023computationally} investigate the utility of low-fidelity approximations in \ac{BO} using three strategies: building a surrogate model for the system dynamics, simplifying the \ac{MPC} policy via a shortened prediction horizon, and coarsening the time grid to complete the simulation in fewer time steps.
\cite{lu2021bayesian} also use a low-fidelity MPC formulation that preserves prior knowledge on the overall structure of the objective function to provide a reference model, which can be utilized in \ac{BO} to reduce the number of queries to the system.

\ac{BO} can also benefit from considering the effect of decisions on those in the future, aiming to acquire informative samples that mitigate the number of future queries.
Casting the \ac{BO} problem into \iac{MDP} formulation, stochastic \ac{DP} can recover a multi-step optimal \ac{AF} that would maximize expected improvement over the remaining query budget in theory \cite{frazier2018tutorial}.
While acquiring such an \ac{AF} is obstructed by the curse of dimensionality \cite{bellman1962applied}, approximations of this multi-step strategy have still demonstrated improved efficiency relative to myopic strategies \cite{jiang2020binoculars}.
One approach leverages rollouts of some base policy as an approximation for the \ac{DP} solution \cite{yue2020why,paulson2022efficient}.
Alternatively, this planning can be cached within a policy or $Q$-function approximator.
These approximators in an \ac{MDP} formulation of \ac{BO} can be learned using model-free \ac{RL} and incorporated into the \ac{AF} towards non-myopic acquisition \cite{volpp2020meta-learning,hsieh2021reinforced}.
Integrating model-free \ac{RL} with \ac{BO} in this fashion has demonstrated success in general single and multi-objective optimization problems and transfer learning \cite{volpp2020meta-learning,hung2025boformer}, but these ideas remain largely unexplored as applied to policy optimization.
       
\section{Policy Gradient \ac{RL}} \label{sec: 5}

This section presents policy gradient \ac{RL} as a gradient-based approach to the policy optimization problem in \cref{eqn: policy search}.
We first introduce its formulation and sample-based estimation of $\nabla_{\theta}J(\pi_{\theta})$.
We then discuss the strengths of policy gradient \ac{RL} for model-free learning of model-based policies, covering both standalone policy gradient methods and actor-critic algorithms, followed by open research opportunities.

\subsection{Algorithmic Components of Policy Gradient \ac{RL}}

\begin{figure}[htb]
\centering
\includegraphics[width=0.9\linewidth]{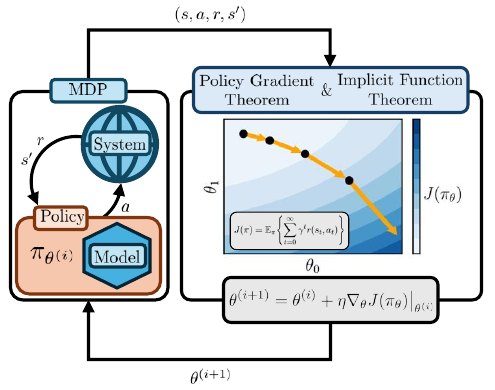}
\caption{
Policy gradient \ac{RL} for model-based policies exploits the \ac{MDP} structure and consequently that of the objective $J(\pi_{\theta})$ to estimate the gradient $\nabla_{\theta}J(\pi_{\theta})$.
Within the \ac{MDP}, a parametric policy $\pi_{\theta}$ interacts with the system to produce $(s,a,r,s')$ tuples.
Using the \ac{PGT}, the gradient $\nabla_{\theta}J(\pi_{\theta})$ (orange) is estimated from these tuples to ascend $J(\pi_{\theta})$ (blue) through iterative updates to parameters $\theta$, which need not be episodic.
For implicitly defined model-based policies, such as \ac{MPC}, the application of the \ac{PGT} is enabled by the \ac{IFT}.
}
\label{fig: policy gradient}
\end{figure}

While \ac{DFO} generally assumes little to no knowledge of $J$, gradient-based methods exploit its structure as an expected cumulative reward under Markovian dynamics to estimate $\nabla_\theta J(\pi_\theta)$ from samples.
The policy parameters are then iteratively updated along these estimates until convergence to a (locally) optimal policy \cite{sutton1999policy}.
A visual representation of this gradient-based policy optimization process can be seen in \Cref{fig: policy gradient}.
Policy gradient methods thus comprise two key ingredients: a \emph{policy gradient estimator} and an \emph{update rule}.

\textbf{Policy gradient estimators.}\quad 
The \ac{PGT} expresses the gradient of \cref{eqn: performance function} in a form that can be estimated from samples, without knowledge of the system dynamics.
\begin{theorem}[\Acl{PGT} \cite{sutton1999policy}]
    For a given \ac{MDP} with state space $\mathcal{S}$ and action space $\mathcal{A}$, let $\pi_{\theta} : \mathcal{S} \rightarrow \Delta(\mathcal{A})$ be a stochastic policy with parameters $\theta$ and density $\pi_{\theta}(a|s)$ differentiable in $\theta$, where $\Delta(\mathcal{A})$ is the set of probability distributions over $\mathcal{A}$.
    Then, the gradient of the performance function $J(\pi_{\theta})$ with respect to $\theta$ satisfies
    \begin{equation} \label{eqn: PGT}
        \nabla_{\theta}J(\pi_{\theta}) \propto \mathbb{E}_{s \sim d^{\pi}\!,\,a \sim \pi_\theta} \bigl\{ \nabla_{\theta}\log\pi_{\theta}(a|s)\, Q^{\pi}(s,a) \bigr\},
    \end{equation}
    where $d^{\pi}$ is the normalized discounted state-visitation distribution under $\pi_\theta$, $Q^{\pi}(s,a)$ is the expected discounted return from state $s$ and action $a$ under $\pi_\theta$, and the proportionality constant is $1/(1-\gamma)$; for brevity, we write $\pi$ for $\pi_\theta$ in superscripts.
\label{thm: PGT}
\end{theorem}
Notably, \cref{eqn: PGT} does not require derivatives of the reward function or of the distribution of visited states, only those of the policy, making the \ac{PGT} suitable for a model-free \ac{RL} setting.
Additionally, for any function $b(s)$ of the state alone, $\mathbb{E}_{a\sim\pi_\theta}\{\nabla_\theta \log \pi_\theta(a|s)\, b(s)\} = 0$ \cite{williams1992simple}, so $Q^\pi$ in \cref{eqn: PGT} may be replaced by $Q^\pi - b$ without biasing the gradient.
Choosing $b = V^\pi$ yields the advantage form
\[
\nabla_\theta J(\pi_\theta) \propto \mathbb{E}_{s \sim d^{\pi}\!,\,a \sim \pi_\theta} \{\nabla_\theta \log \pi_\theta(a|s)\, A^{\pi}(s,a)\},
\]
where $A^\pi(s,a) = Q^\pi(s,a) - V^\pi(s)$ quantifies how much better action $a$ is than the policy's average behavior at $s$.

Policy gradient methods differ chiefly in how they estimate $A^\pi$, trading bias for variance.
Monte Carlo methods use the sampled returns $G_t = \sum_{k=t}^{T}\gamma^{k-t} r_k$ for finite trajectories of length $T$, optionally minus a learned baseline $V_\phi(s_t)$, yielding unbiased but high-variance estimates that require complete episodes \cite{williams1992simple}.
Actor-critic methods instead use a learned critic, e.g., $Q_\phi(s,a) - V_\phi(s)$ or the \ac{TD} error $r + \gamma V_\phi(s') - V_\phi(s)$, which reduces variance and enables per-step updates in continuing tasks, at the cost of bias from critic errors \cite{grondman2012survey}.
Generalized advantage estimation interpolates between these extremes \cite{schulman2018high-dimensional}.

\Cref{thm: PGT} considers a stochastic policy, and thus does not readily accommodate model-based policies such as \cref{eqn: MPC policy}.
In practice, \cref{eqn: MPC policy} can be made stochastic by including a random variable as policy input \cite{gros2021reinforcement} or adding a small amount of noise to computed actions \cite{banker2025gradient-based}, e.g.,
\[
\pi_\theta(a|s) = \mathcal{N}(u_{0|t}^\star,\Sigma),
\]
for which $\nabla_\theta \log \pi_\theta(a|s) = \big(\nabla_\theta u_{0|t}^\star\big)^{\top}\Sigma^{-1}\big(a - u_{0|t}^\star\big)$, with $\nabla_\theta u_{0|t}^\star$ given by \cref{eqn: MPC sensitivity}.
An alternative that avoids this conversion is the deterministic \ac{PGT} \cite{silver2014deterministic},
\[
\nabla_{\theta}J(\pi_{\theta}) \propto \mathbb{E}_{s \sim d^{\pi}} \bigl\{ \nabla_{\theta}\pi_{\theta}(s)^{\top} \nabla_{a} Q^{\pi}(s,a)\big|_{a=\pi_{\theta}(s)} \bigr\},
\]
which admits \ac{MPC} policies directly, with $\nabla_\theta \pi_\theta(s)$ given by \cref{eqn: MPC sensitivity}.
Because $\nabla_a Q^{\pi}$ must be estimated by a learned critic, the deterministic \ac{PGT} leads naturally to actor-critic methods \cite{gros2020data-driven}.
By avoiding the integral over actions, it can also reduce the samples required in large action spaces and permit off-policy learning without importance weighting, enabling sample reuse \cite{silver2014deterministic,lillicrap2016continuous}.

\textbf{Update rule.}\quad After selecting a gradient estimator, the policy parameters are iteratively updated according to a complementary update rule, e.g.
\begin{equation} \label{eqn: gradient step}
    \theta^{(i+1)} = \theta^{(i)} + \eta \nabla_{\theta}J(\pi_{\theta})\big|_{\theta^{(i)}},
\end{equation}
where $\eta$ is a positive scalar step size or, more generally, a positive-definite matrix.
In actor-critic algorithms, these policy updates are interleaved with critic updates so that the critic tracks the changing policy, keeping gradient estimates accurate.
While \cref{eqn: gradient step} is a basic example, more sophisticated rules augment it with preconditioning or constraints for efficient and safe policy optimization.

Preconditioned updates choose $\eta$ as a matrix that accounts for the local curvature and scaling of the performance function or of the policy parameterization.
(Quasi-)Newton methods address the former, using the curvature of $J$ itself \cite{brandner2024reinforced,kordabad2026quasi-newton}.
Such second-order updates are often impractical for \ac{NN} policies with millions of parameters, but are affordable for model-based policies, whose few parameters are often heterogeneous in scale, e.g., cost weights alongside model coefficients.
Natural gradient methods address the latter, using the inverse Fisher information matrix, which measures how the policy's action distribution changes with $\theta$, to obtain the steepest ascent direction under a local KL-divergence constraint between successive policies \cite{kakade2001natural}.

Constrained updates instead restrict where each step can go.
Trust-region methods bound the change in the policy at each step, either explicitly \cite{schulman2015trust} or approximately via clipping \cite{schulman2017proximal}, and safe \ac{RL} methods further constrain updates to satisfy cost or Lyapunov conditions \cite{achiam2017constrained,chow2019lyapunov}.
For model-based policies, such constraints can be imposed directly on $\theta$, e.g., to preserve the feasibility, stability, and safety of the \ac{MPC} scheme throughout learning \cite{gros2022learning,gros2019safeii}.

\subsection{Strengths for Control Policy Optimization}

Two strengths make policy gradient \ac{RL} particularly suited to control policy optimization: the \emph{diverse learning schemes} it supports, and its \emph{scalability} with respect to both parameters and data.

\textbf{Diverse learning schemes.}\quad
When experience is cheap to generate and rewards are dense, Monte Carlo policy gradient methods such as REINFORCE can be effective for policy optimization \cite{sutton2018reinforcement}, but they are restricted to \emph{episodic}, \emph{on-policy} learning: episodes must have finite length, and samples must be generated under the current policy to approximate $Q^{\pi}$ in \cref{eqn: PGT}.
Numerous policy gradient schemes have been developed to address these restrictions, enabling model-based policies to learn \emph{non-episodically}, from individual state transitions with immediate policy updates, and \emph{off-policy}, from interactions generated by policies other than the one being learned.

\emph{Non-episodic learning:}\quad 
The gradient estimators and update rules of policy gradient \ac{RL} utilize information at the level of individual transitions $(s,a,r,s')$, not only scalar observations of $J$.
Whereas \ac{DFO} assigns a single performance value to each tested parameter, transition-level data reveal how each decision contributed to performance: the \ac{PGT} exploits the Markov structure of the \ac{MDP} to attribute returns to individual actions, and learned critics further exploit the Bellman equation to propagate this credit across time.
For model-based policies, the \ac{IFT} additionally exploits the structure of \cref{eqn: mpc problem} to translate this credit into parameter updates via \cref{eqn: MPC sensitivity}.

A major benefit of this is that the policy can be learned during its online deployment.
As state transitions are observed, the policy can be immediately updated toward improving its infinite-horizon performance.
Provided the closed-loop system remains stable throughout learning, these updates can be done indefinitely without a predefined termination condition for the episode.
While such assumptions are difficult to ensure for generic \ac{NN} policies, the control-theoretic structure of \cref{eqn: mpc problem} can make it enforceable.
For example, \cite{gros2019safeii} update robust \ac{MPC} policies via deterministic policy gradients while constraining the updates to preserve safety, and \cite{gros2022learning} establish conditions under which parameter updates preserve the stability and safety of \ac{MPC}-based policies.

\emph{Off-policy learning:}\quad 
The gradient estimators introduced previously can be computed in an on-policy or off-policy manner.
On-policy approaches estimate the advantage by rolling out the current policy, and these rollouts are typically used for a single update and then discarded, which is inefficient when data are expensive.
Off-policy methods instead learn from interactions generated by other behavior policies, e.g., earlier iterates, enabling experience replay and its prioritized variants to reuse each interaction across many updates \cite{lin1992self-improving,schaul2016prioritized}.
For stochastic policies, off-policy gradient estimates generally require importance weighting, which can inflate variance; the deterministic \ac{PGT} avoids this.

Accordingly, off-policy learning of model-based policies has largely relied on the deterministic \ac{PGT}.
\cite{cai2021mpc-based} apply deterministic policy gradients to learn the cost, dynamics, and constraints of an \ac{MPC} policy.
\cite{kordabad2021mpc-based} extend deterministic policy gradients with advantage estimation to learn ``bang-bang'' optimal policies, whose sensitivities tend to zero for much of the state space.
\cite{anand2023painless} eliminate the need for a separate critic network for estimating deterministic policy gradients, simplifying the actor-critic algorithm by leveraging the terminal value function of \ac{MPC}.
The structure of model-based policies suggests opportunities beyond these schemes.
For example, since references or goals enter \cref{eqn: mpc problem} explicitly, hindsight relabeling of goals \cite{andrychowicz2017hindsight} could naturally extend to \ac{MPC} policies \cite{lawrence2025view}.

\textbf{Scalability in learning.}\quad Policy gradient methods exploit the known structure of \cref{eqn: performance function} to estimate its gradient, and can draw on concepts from \ac{DP} to build structured learning schemes such as actor-critic algorithms.
As a result, policy gradient \ac{RL} scales favorably with the number of policy \emph{parameters}, owing to gradient-based optimization, and with the amount of \emph{data}, since structured learning schemes improve gradient estimation.

\emph{Parameters:}\quad 
As discussed in \Cref{sec: 4}, learning high-dimensional policy parameterizations is challenging since the volume of the search space, and hence the number of evaluations needed to explore it, grows exponentially with the number of parameters.
Policy gradient \ac{RL} instead provides a search direction for improving the policy, converging to a local optimum under standard conditions \cite{sutton2018reinforcement} and to a globally optimal policy in select settings \cite{fazel2018global,zhang2020sample,bhandari2022global}.
The update rules discussed above, e.g., adaptive step sizes and trust regions, further stabilize learning in high dimensions \cite{schulman2015trust,schulman2017proximal}.
Moreover, the cost of each gradient estimate scales mildly with the number of parameters.
For \ac{MPC} policies, a single linear solve with the transposed \ac{KKT} matrix yields the gradient with respect to all parameters simultaneously \cite{amos2018differentiable}, whereas each \ac{BO} iteration requires refitting a surrogate of $J(\pi_\theta)$ and optimizing a potentially multimodal \ac{AF} \cite{binois2022survey}.

This strength of policy gradient \ac{RL} is most evident in learning \ac{NN} policies, which policy gradient methods have trained with tens of thousands \cite{schulman2015trust}, hundreds of thousands \cite{lillicrap2016continuous,haarnoja2018soft}, and, most recently, billions of parameters \cite{deepseekai2025deepseekr1}.
For model-based policies, policy gradients have been shown to outperform \ac{DFO} for high-dimensional \ac{MPC} parameterizations while exploiting both the \ac{MDP} and \ac{MPC} structures \cite{banker2025gradient-based}.
This is especially relevant for learning model-based policies where a subset of their components are parameterized by \acp{NN} to yield expressive optimization-based policies.
Examples include learning \ac{NN}-based control objectives \cite{seel2022convex,romero2024actor-critic,jahncke2026differentiable} or dynamics \cite{wan2024difftori} to solve tasks with constrained nonlinear dynamics or high-dimensional observations.

\emph{Data:}\quad 
Policy gradient estimation, and ultimately closed-loop performance, generally improve with access to more data covering the state-action space.
This can be attributed to reducing the variance of $\nabla_{\theta}J(\pi_{\theta})$ estimates, or to improved approximation of $Q^{\pi}(s,a)$ in \cref{eqn: PGT} for actor-critic algorithms.
Actor-critic algorithms are particularly well suited to large datasets because the Bellman equation imposes structure on how future performance is represented.
Rather than modeling $J(\pi_\theta)$ directly, the critic models the value of individual states and actions, constrained by the recursive relationship that each value equals the immediate reward plus the discounted value of the successor state.
Every transition $(s,a,r,s')$ therefore provides a consistency condition on the critic through bootstrapping, and transitions collected under different policies can inform the same critic via off-policy experience replay.
In contrast, a \ac{GP} surrogate in \ac{BO} assumes little about $J(\pi_\theta)$ beyond the regularity encoded by its kernel, and receives only a single scalar observation per evaluated policy, so it cannot exploit the temporal decomposition of performance.
Moreover, $Q_{\phi}(s,a)$ may be approximated by \iac{NN}, or other model classes whose training and inference scale favorably with the amount of data.
Both of these are challenges for \ac{BO} methods, especially in high dimensions \cite{eriksson2019scalable}.

These benefits are most readily observed for \ac{NN} policies due to the large body of existing work, but recent work shows that model-based policies can similarly benefit.
\cite{schulman2015trust} demonstrate the scalability of policy gradient \ac{RL} relative to \ac{DFO} methods such as the cross-entropy method and covariance matrix adaptation, using thousands of transitions per policy update.
\cite{lillicrap2016continuous} demonstrate how experience replay in actor-critic algorithms can improve the sample efficiency of policy gradient \ac{RL}, learning \ac{NN} policies from a history of $10^6$ tuples.
For model-based policies, \cite{seel2022convex} learn a convex control objective from thousands of individual transitions, while \cite{romero2024actor-critic,wan2024difftori} scale to millions of samples for learning \ac{MPC} policies.
These examples, while few, indicate significant opportunities for learning model-based policies from large amounts of data, particularly with actor-critic algorithms.

\subsection{Opportunities for Future Work} \label{sec: 5.3}
As discussed in \Cref{sec: 3}, model-based policies offer benefits in safety and modular structure.
Fully realizing these benefits in policy gradient \ac{RL} calls for two promising research directions: developing \emph{control-theoretic learning schemes} that exploit the modularity of model-based policies and incorporate control-theoretic methods into policy gradient \ac{RL}, and integrating \emph{state-of-the-art \ac{RL}} into the model-based policy paradigm while reducing implementation barriers.

\textbf{Control-theoretic learning schemes.}\quad Due to the modular structure of a model-based policy, a subset of individual components---$f,g,h,\ell,$ and $V$---can be learned independently.
This, in turn, provides the flexibility to decompose the policy optimization problem into smaller subproblems, wherein control-theoretic methods can be used to provide theoretical guarantees related to stability, feasibility, and constraint satisfaction of a model-based policy.
As examples, the design of a model-based policy can involve:
\begin{enumerate}[-]
    \item learning a \emph{control Lyapunov function} within the control objective to ensure closed-loop stability \cite{mayne2000constrained};
    \item learning a \emph{control barrier function} and guaranteeing constraint satisfaction for safety \cite{ames2019control};
    \item \emph{uncertainty quantification} in system identification to ensure robust constraint satisfaction \cite{lucia2013multi-stage}; or
    \item following the \emph{notion of \ac{I4C}} to obtain performance-oriented models \cite{gevers1993towards}.
\end{enumerate}

Regarding system identification, \cite{banker2025gradient-based} and \cite{martinsen2020combining} provide two approaches for learning the dynamics model via \ac{RL}.
\cite{banker2025gradient-based} rely on the notion of \ac{I4C} while \cite{martinsen2020combining} aim to manage the conflicting state prediction and \ac{RL} objectives.
\cite{gros2020safe} and \cite{gros2022learning} provide strategies for safe policy gradient \ac{RL} for \ac{MPC} in systems with linear dynamics.
\cite{gros2020safe} project actions to a safe set via robust \ac{MPC} to ensure safety during policy optimization.
\cite{gros2022learning} establish conditions under which online parameter updates preserve the safety and stability of the \ac{MPC} policy.
\cite{banker2026stability} present a policy gradient \ac{RL} approach that decouples the stability requirement from the global performance function.
Going beyond quadratic control objectives, \cite{seel2022convex} learn \iac{NN}-based cost modification that can guarantee nominal stability.
Application of these guarantees, however, has largely been restricted to linear systems or nominal \ac{MPC}.
Important open directions include extending control-theoretic guarantees to nonlinear, stochastic, and partially observed systems; maintaining them during learning and exploration, not only at convergence; and systematically deciding which components should be learned for performance and which for predictive accuracy.
Challenging benchmarks \cite{gymnasium_robotics2023github,towers2024gymnasium} offer a means to evaluate such schemes at scale.

\textbf{State-of-the-art \ac{RL}.}\quad
Investigations of model-free \ac{RL} of model-based policies have generally considered simple \ac{RL} schemes, such as REINFORCE \cite{williams1992simple} and deterministic policy gradient algorithms \cite{silver2014deterministic}.
While these classical algorithms are useful, learning model-based policies can benefit from broader advancements in \ac{RL}, including:
\begin{enumerate}[-]
    \item \emph{goal-conditioned \ac{RL}} to learn model-based policies that can generalize across multiple goals, including under sparse goal-reaching rewards \cite{andrychowicz2017hindsight,chane-sane2021goal-conditioned};
    \item \emph{intrinsic motivation} \cite{chentanez2004intrinsically} or \emph{reward shaping} \cite{ng1999policy} to aid a model-based policy in exploring the system and incrementally learning its components when external rewards are sparse or even absent;
    \item \emph{\ac{RL} from human feedback} for human-in-the-loop policy optimization that aligns a model-based policy with human preferences \cite{christiano2017deep,ouyang2022training};
    \item \emph{\ac{RL} as probabilistic inference} \cite{levine2018reinforcement,haarnoja2018soft} and \emph{distributional \ac{RL}} \cite{bellemare2017distributional} for uncertainty estimation to enhance exploration, or in robust or stochastic control strategies that rely on uncertainty estimates;
    \item \emph{meta \ac{RL}} \cite{nagabandi2019learning} and \emph{sim2real transfer} \cite{tan2018sim-to-real} for efficient online adaptation of the various components comprising model-based policies in dynamic systems;
    \item \emph{multi-agent \ac{RL}} to learn model-based policies with coordinated behaviors \cite{tan1993multi-agent}, or \emph{hierarchical \ac{RL}} for coordinating model-based policies whose components and decisions correspond to different levels of temporal and behavioral abstraction \cite{nachum2018data-efficient}; and
    \item \emph{partially observable \ac{RL}} where a model-based policy must design actions without complete knowledge of the state of the system \cite{krishnamurthy2016partially}.
\end{enumerate}

Exploring these \ac{RL} strategies with model-based policies would benefit from lower implementation and computational barriers to advanced \ac{RL} schemes.
Currently, integrating \ac{MPC} with \ac{RL} can be cumbersome and expensive, both because the two fields' highly specialized tools must be implemented within a single ecosystem, and because running and differentiating \ac{MPC} repeatedly within iterative \ac{RL} schemes is computationally costly.
Several frameworks have emerged to reduce these barriers.
One of the earliest examples for bridging the two ecosystems is \cite{amos2018differentiable}, who develop a differentiable \ac{MPC} policy for easy implementation in an \ac{RL} ecosystem.
\cite{adabag2026differentiable} also develop a differentiable \ac{MPC} policy designed for GPU acceleration.
\cite{salzmann2024learning} ease integration from the opposite direction with a framework for integrating machine learning models, potentially learned via \ac{RL}, into an \ac{MPC} ecosystem.
\cite{frey2025differentiable} implement \ac{RL} algorithms within an \ac{MPC} ecosystem with the added benefit of efficient computation of \ac{MPC} derivatives.
\cite{lawrence2025mpcritic} develop a model-based policy learning framework, which facilitates easy transfer of models between \ac{RL} and \ac{MPC} ecosystems and sidesteps differentiating the \ac{MPC} policy.
Continued development of such frameworks can aid in accommodating state-of-the-art \ac{RL}, especially for sophisticated \ac{MPC} formulations.

\section{Offline Learning} \label{sec: 6}

Policy optimization with \ac{BO} or policy gradient \ac{RL} assumes the ability to interact with the system, but this can be infeasible due to cost and/or safety considerations, as described in \Cref{sec: 3}.
When unable to collect new experience, the policy must be learned from a fixed, offline dataset.
A shared challenge among strategies that learn from an offline dataset is distribution shift \cite{levine2020offline}: the offline dataset models the distribution of states $d^{\pi_{\beta}}(s)$ induced by some behavior policy $\pi_{\beta}$, whereas evaluating the learned policy during online deployment is apt to induce another distribution $d^{\pi_{\theta}}(s)$.
Thus, direct application of \cref{eqn: PGT} under $d^{\pi_{\beta}}(s)$ can yield biased estimates for $\nabla_{\theta}J(\pi_{\theta})$.
To extract an effective policy, learning policies offline must consider whether the offline dataset is generated from expert or suboptimal policies, if reward information is available, and how to mitigate distribution shift.
We briefly describe \acf{BC} and offline \ac{RL} as distinct strategies in light of this.

\textbf{Behavioral cloning.}\quad Given an offline dataset of $(s,a)$ tuples generated by an expert policy, \ac{BC} aims to learn a policy that best replicates the limited expert data \cite{pomerleau1991efficient}.
For model-based policies, this can manifest as learning the dynamics model \cite{amos2018differentiable}, cost function \cite{jin2020pontryagin}, or constraints that result in expert-like decisions.
However, these early examples do not address the challenge of generalizing beyond the demonstrations.
The dataset can be expanded by deploying the learned policy and querying the expert on visited states \cite{ross2011reduction}, but this requires an expert available during deployment.
Instead, the structure of a model-based policy is well poised to enforce constraints that limit distribution shift, without necessarily gathering more data, as discussed below.

\textbf{Offline \ac{RL}.}\quad Given an offline dataset of $(s,a,r,s')$ tuples, possibly generated under a non-expert policy, offline \ac{RL} aims to extract a policy that maximizes \cref{eqn: performance function} \cite{wu2019behavior}.
Since most offline \ac{RL} methods are actor-critic, performance depends primarily on (i) value or $Q$-function estimation; (ii) extraction of a policy from the learned values; and (iii) generalization of the policy for out-of-distribution states encountered online \cite{park2024value}.
While offline \ac{RL} of \ac{NN} policies has received significant attention \cite{levine2020offline}, offline \ac{RL} of model-based policies has been little explored.
\cite{sawant2023learning-based} is one of the few works in this space, presenting a value-based method that learns \ac{MPC} parameters from large offline datasets without solving the \ac{MPC} problem over the data.
Nonetheless, the critical challenges of offline \ac{RL} \cite{levine2020offline,park2024value}, including divergence in value estimation and limited policy improvement when the data are suboptimal, remain largely unexplored.

Both \ac{BC} and offline \ac{RL} suffer from distribution shift: in \ac{BC}, small imitation errors compound into states absent from the demonstrations \cite{ross2011reduction}, while in offline \ac{RL}, out-of-distribution actions corrupt Bellman targets, e.g., $q$ in \cref{eqn: mean squared bellman error minimization}, causing value estimates to diverge, and the deployed policy may drive the system away from the data \cite{levine2020offline}.
Model-based policies offer structural remedies to both.
Constraints can be embedded directly in the policy rather than balanced as soft penalties \cite{yu2020mopo}, e.g., \cite{acerbo2023evaluation} enforce lane constraints over a finite horizon to avoid states unseen in training while learning a cost for human-like driving.
Model-based policies can extend such reasoning to infinite-horizon guarantees of remaining in distribution using approaches based on Lyapunov density models \cite{kang2022lyapunov} or control barrier functions \cite{castaneda2023distribution}.
Moreover, uncertainty estimates for dynamics models can enable robust or stochastic \ac{MPC} formulations that mitigate performance degradation under distribution shift.
Finally, the two settings are complementary: while \ac{BC} generally cannot exceed the performance of the demonstrator, it can initialize model-based policies that are subsequently improved with \ac{RL} \cite{chen2019gnu-rl}.
Although nascent, research integrating control-theoretic structures into these settings holds potential for obtaining safe, high-performing policies without online interaction.

\section{Concluding Remarks} \label{sec: 7}     

Policy optimization is a versatile and scalable approach towards obtaining approximately optimal decision-making under uncertainty.
Within this framework, model-based policies provide structure for efficiently learning safe and interpretable decision-making policies in a model-free learning setting.
Whereas \acp{NN} are widely used to approximate the policy and $Q$-function, a model-based policy can employ existing models of the cost, dynamics, and constraints to approximate its policy or $Q$-function, and these approximations can be refined by model-free learning.
In contrast to expressive but data-intensive \ac{NN} policies, a model-based policy embodies an inductive bias for sample-efficient learning, while still being able to approach the asymptotic performance of model-free learning.
A prototypical example of model-based policies is \ac{MPC} due to its interpretable, safe, and modular formulation.
Model-based policies can be learned through either \ac{DFO} or gradient-based methods; the assumptions and tools behind each lead to distinct strengths and challenges.

In this paper, we discussed \ac{BO}, policy gradient \ac{RL}, and \ac{BC} as approaches for learning a model-based policy, summarizing their strengths in \Cref{tab: strengths}.
\ac{BO} is a flexible \ac{DFO} approach to policy optimization, which can effectively balance trade-offs, but largely discards the \ac{MDP} problem structure.
Policy gradient \ac{RL} instead exploits this structure, incorporating elements of \ac{DP} in certain settings, to obtain diverse learning schemes that scale with the learnable parameters and data, at the cost of differentiating the policy.
\ac{BC} is a pragmatic approach when expert decisions are available but only indirectly targets performance on the true system.
Across these approaches, recurring opportunities include preserving control-theoretic guarantees throughout learning, scaling the computation and differentiation of optimization-based policies, and extending beyond standard \acp{MDP}, as discussed in \Cref{sec: 4.3,sec: 5.3,sec: 6}.

A promising direction, which we have only touched on, is formulating a model-based policy in terms of an approximate $Q$-function.
Like policy approximation, $Q$-function approximation can embed prior (potentially physics-based) models of the system, enabling sample-efficient learning of model-based policies while maintaining a stronger connection to \ac{DP}.
Taking a local-global perspective \cite{banker2025local-global}, the \ac{MPC} problem in \cref{eqn: mpc problem} can be viewed as a $Q$-function approximator.
Globally, the approximator is updated to approximately satisfy the Bellman equation for all states; locally, the model-based policy acts by taking the action that minimizes this approximator at each state.
Informed and constrained with prior knowledge of the system, this local optimization provides a basis for safety and stability guarantees through existing control-theoretic methods.
In turn, the resulting local interactions drive improvements of the $Q$-function approximator towards satisfying the global Bellman equation, improving the policy's decisions.
Hence, incorporating model-based policies into a local-global framework is a rich area for future research, investigating the interplay of value function approximation and model-based control.

    \begin{table*}[t]
        \centering
        \caption{Comparison of model-free learning methods for model-based policies.
        For each aspect, we denote each method as generally strong/mature (\checkmark), promising/nascent (+), unexplored ($\ocircle$), weak overall ($\times$), or not applicable ($-$), while providing a non-exhaustive list of references.}
        \begin{tabular}{ccrlrlrl}
            \toprule
                 \multicolumn{2}{c}{Aspect} & \multicolumn{2}{l}{\textbf{Bayesian Optimization}} & \multicolumn{2}{l}{\textbf{Policy Gradient \ac{RL}}} & \multicolumn{2}{l}{\textbf{Behavioral Cloning}} \\
            \midrule 
                \multirow{2}{*}{\textbf{Trade-offs}} & Multi-objective & \checkmark & \cite{gharib2021multi-objective,makrygiorgos2022performance-oriented,chan2024practical,hoang2025multi-objective} & $\ocircle$ & & $-$ & \\
                & Safety or Constraints & \checkmark & \cite{sorourifar2021data,lu2022no-regret,hirt2024stability-informed,berkenkamp2023bayesian,krishnamoorthy2023model-free,chan2023safe} & + & \cite{gros2019safeii,gros2020safe,gros2022learning,banker2026stability} & + & \cite{east2020infinite-horizon,jin2021safe,oshin2023differentiable} \\
            \cmidrule(rl){1-2}
                \multirow{2}{*}{\textbf{Flexibility}} & Policy Parameterization & \checkmark & \cite{piga2019performance-oriented,choksi2021simulation-based} & + & 
                \cite{gros2020reinforcement,kordabad2021reinforcement,seel2022convex} & + & \cite{amos2018differentiable,jin2020pontryagin} \\
                & Beyond Standard \acp{MDP} & \checkmark & \cite{brunzema2022on,cho2024run-indexed,shao2024coactive,zhu2021preference-based,shao2024time-varying} & $+$ & \cite{esfahani2022policy} & $\ocircle$ & \\
            \cmidrule(rl){1-2}
                \multirow{2}{*}{\textbf{Scalability}} & Parameters & + & \cite{kudva2024efficient,frohlich2019bayesian,frohlich2021cautious} & \checkmark & \cite{romero2024actor-critic,banker2025gradient-based,jahncke2026differentiable} & \checkmark & \cite{xiao2023learning,acerbo2023evaluation,acerbo2024driving} \\
                & Data & $\ocircle$ & & \checkmark & \cite{wan2024difftori,frey2025differentiable,bravo2026turbompc} & \checkmark & \cite{wan2024difftori,frey2025differentiable,bravo2026turbompc} \\
            \cmidrule(rl){1-2}
                \multirow{3}{*}{\textbf{Learning}} & Non-episodic & $\times$ & & \checkmark & 
                \cite{gros2021reinforcement,kordabad2021mpc-based} & $-$ & \\
                & Off-policy & $-$ & & \checkmark & \cite{cai2021mpc-based,anand2023painless} & $-$ & \\
                & Offline & $\times$ & & $\ocircle$ & & \checkmark & \cite{chen2019gnu-rl,rickenbach2025zipmpc} \\
            \bottomrule
        \end{tabular}
        \label{tab: strengths}
    \end{table*}

\begin{ack}                               
  The authors thank Nathan P. Lawrence for insightful discussions. This work was supported in part by the U.S. National Science Foundation under grant 2317629.
\end{ack}

\bibliographystyle{plain}        
\bibliography{references}           

\end{document}

%% file: acronyms.tex
\DeclareAcronym{IFT}{short = IFT,
        long = Dini's implicit function theorem}

\DeclareAcronym{SVD}{short = SVD,
        long = singular value decomposition}
\DeclareAcronym{PCA}{short = PCA,
	long = principal component analysis}
\DeclareAcronym{RMSE}{short = RMSE,
	long = root mean square error}	

\DeclareAcronym{KKT}{short = KKT,
	long = Karush-Kuhn-Tucker}
\DeclareAcronym{LICQ}{short = LICQ,
	long = linear independence constraint qualification}
\DeclareAcronym{SOSC}{short = SOSC,
	long = second-order sufficient conditions of optimality}
\DeclareAcronym{DFO}{short = DFO,
        long = derivative-free optimization}
\DeclareAcronym{MO}{short = MO,
	long = multi-objective,
    short-indefinite = an}
\DeclareAcronym{NLP}{short = NLP,
	long = nonlinear programming}

\DeclareAcronym{AI}{short = AI,
	long = artificial intelligence}
\DeclareAcronym{ML}{short = ML,
	long = machine learning,
	short-indefinite = an}
\DeclareAcronym{AD}{short = AD,
	long = automatic differentiation}	
\DeclareAcronym{NN}{short = NN,
	long = neural network,
    short-indefinite = an}	
\DeclareAcronym{BN}{short = BN,
	long = Bayesian network}	
\DeclareAcronym{CNN}{short = CNN,
	long = convolutional neural network}	
\DeclareAcronym{DNN}{short = DNN,
	long = deep neural network}	
\DeclareAcronym{BNN}{short = BNN,
	long = Bayesian neural network}	
\DeclareAcronym{MLP}{short = MLP,
	long = multilayer perceptron,
	short-indefinite = an}
\DeclareAcronym{TL}{short = TL,
	long = transfer learning}	
\DeclareAcronym{VAE}{short = VAE,
	long = variational autoencoder}	

\DeclareAcronym{BO}{short = BO,
	long = Bayesian optimization}
\DeclareAcronym{AF}{short = AF,
	long = acquisition function}
\DeclareAcronym{GP}{short = GP,
	long = Gaussian process,
	long-plural-form = Gaussian processes}
\DeclareAcronym{RBF}{short = RBF,
	long = radial basis function,
	short-indefinite = an}	
\DeclareAcronym{GPR}{short = GPR,
	long = Gaussian process regression}	

\DeclareAcronym{MPC}{short = MPC, 
	long = model predictive control,
	long-plural-form = model predictive controllers,
	short-indefinite = an}		
\DeclareAcronym{LQR}{short = LQR, 
	long = linear quadratic regulator}
\DeclareAcronym{MPPI}{short = MPPI, 
	long = model predictive path integral,
	long-plural-form = model predictive path integrals,
	short-indefinite = an}		
\DeclareAcronym{DARE}{short = DARE, 
	long = discrete algebraic Riccati equation}
\DeclareAcronym{LTI}{short = LTI, 
	long = linear time-invariant,
	short-indefinite = an}
\DeclareAcronym{I4C}{short = I4C, 
	long = identification for control}

\DeclareAcronym{MDP}{short = MDP,
	long = Markov decision process,
	long-plural-form = Markov decision processes,
	short-indefinite = an}
\DeclareAcronym{DP}{short = DP,
	long = dynamic programming}
\DeclareAcronym{RL}{short = RL, long = reinforcement learning, short-indefinite = an}
\DeclareAcronym{ADP}{short = ADP,
	long = approximate dynamic programming}
\DeclareAcronym{PGT}{short = PGT,
        long = policy gradient theorem}
\DeclareAcronym{DDPG}{short = DDPG,
	long = deep deterministic policy gradient}
\DeclareAcronym{DPG}{short = DPG,
	long = deterministic policy gradient}
\DeclareAcronym{DQN}{short = DQN,
	long = deep $Q$-network}
\DeclareAcronym{HJB}{short = HJB,
	long = {Hamilton-Jacobi-Bellman}}
\DeclareAcronym{PPO}{short = PPO,
	long = proximal policy optimization}
\DeclareAcronym{REINFORCE}{short = REINFORCE,
	long = {\emph{RE}ward Increment $=$ \emph{N}onnegative \emph{F}actor $\times$ \emph{O}ffset \emph{R}einforcement $\times$ \emph{C}haracteristic \emph{E}ligibility}}
\DeclareAcronym{SAC}{short = SAC,
	long = soft actor-critic}	
\DeclareAcronym{TD3}{short = TD3,
	long = twin-delayed DDPG}
\DeclareAcronym{A3C}{short = A3C,
	long = asynchronous advantage actor-critic}	
\DeclareAcronym{HER}{short = HER,
	long = hindsight experience replay}	
\DeclareAcronym{GPS}{short = GPS,
        long = guided policy search}
\DeclareAcronym{IL}{short = IL,
        long = imitation learning}
\DeclareAcronym{BC}{short = BC,
        long = behavioral cloning}
\DeclareAcronym{TD}{short = TD,
        long = temporal difference}
	
\DeclareAcronym{CSTR}{short = CSTR,
	long = continuous stirred tank reactor}	